\documentclass{article}

\usepackage[preprint]{neurips_2026}

\usepackage[utf8]{inputenc}
\usepackage[T1]{fontenc}
\usepackage{hyperref}
\usepackage{url}
\usepackage{booktabs}
\usepackage{amsfonts}
\usepackage{amsmath}
\usepackage{amssymb}
\usepackage{nicefrac}
\usepackage{microtype}
\usepackage{xcolor}
\usepackage{graphicx}
\usepackage{subcaption}
\usepackage{algorithm2e}
\usepackage{makecell}
\usepackage[capitalize]{cleveref}

\title{Soft Geometric Inductive Bias for Object Centric Dynamics}

\author{%
  Hampus Linander\,\textsuperscript{1,$\ast$,$\ddagger$} \qquad
  Conor Heins\,\textsuperscript{2,$\ddagger$} \qquad
  Alexander Tschantz\,\textsuperscript{3,$\ddagger$} \\
  \bfseries Marco Perin\,\textsuperscript{4,$\ddagger$} \qquad
  Christopher L Buckley\,\textsuperscript{3,$\ddagger$} \AND
  \normalfont\small
  \begin{tabular}[t]{@{}l@{}}
    \begin{tabular}[t]{@{}r@{}l@{}}
      \textsuperscript{1} & Department of Mathematical Sciences \\
                          & Chalmers University of Technology \\
                          & University of Gothenburg
    \end{tabular} \\ \noalign{\vskip 8pt}
    \begin{tabular}[t]{@{}r@{}l@{}}
      \textsuperscript{3} & Department of Informatics \\
                          & University of Sussex
    \end{tabular}
  \end{tabular}\hspace{2.5em}
  \begin{tabular}[t]{@{}l@{}}
    \begin{tabular}[t]{@{}r@{}l@{}}
      \textsuperscript{2} & Department of Collective Behavior \\
                          & Max Planck Institute of Animal Behavior
    \end{tabular} \\ \noalign{\vskip 8pt}
    \begin{tabular}[t]{@{}r@{}l@{}}
      \textsuperscript{4} & Kodamai
    \end{tabular}
  \end{tabular}
}

\begin{document}
\maketitle
{\renewcommand{\thefootnote}{$\ast$}\footnotetext{Correspondence to: \texttt{linander@chalmers.se}}}
{\renewcommand{\thefootnote}{$\ddagger$}\footnotetext{Part of this work was carried out at VERSES AI.}}
\begin{abstract}
Physical systems are naturally parameterized in terms of geometric entities and their transformations. While exact equivariance can be a powerful inductive bias, real-world applications frequently exhibit approximate or broken symmetries in practice. We introduce object-centric world models that provide a soft geometric inductive bias without enforcing exact symmetries. By embedding object states as Clifford multivectors, our models encourage physically meaningful transformations while retaining the expressivity needed to handle asymmetric dynamics. We evaluate our approach on 2D rigid-body dynamics, 3D charged-particle systems, and real-world driving trajectories. Compared to both unstructured baselines and strictly equivariant models, our soft Clifford transformer achieves better long-horizon fidelity, particularly in regimes with broken symmetries. These results suggest that geometric algebra offers an effective middle ground, delivering sample-efficient dynamics without the inflexibility of hard mathematical constraints.
\\[4pt]{\small Code: \url{https://github.com/hlinander/soft-geometric-inductive-bias}}
\end{abstract}
\begin{figure}[h]
    \centering
    \includegraphics[width=0.9\linewidth]{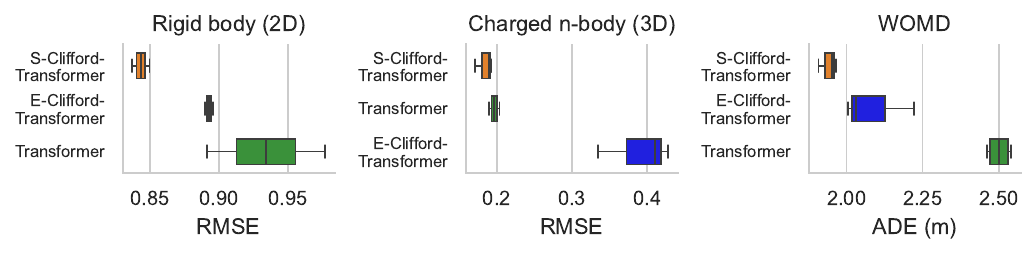}
    \caption{Rollout RMSE and average displacement error (ADE) for autoregressive object-centric world models trained
    on 2D rigid body physics with collisions (left), 3D charged particles in
    harmonic potential (middle), and real-world multi-agent driving scenarios from the Waymo Open Motion Dataset (WOMD)
    (right). Our Clifford transformer with a soft geometric inductive bias (\texttt{S-CliffordTransformer}) outperforms the vanilla transformer (\texttt{Transformer}) and the equivariant Clifford transformer (\texttt{E-CliffordTransformer}) in all data domains.} \label{fig:first_page_combined}
\end{figure}

\section{Introduction}
\label{sec:intro}

Learning world models that generalize far beyond their training distribution remains a central goal in machine learning, especially for applications in computer vision and robotics. Architectural priors---for example, weight sharing or equivariance to geometric transformations---can dramatically improve data efficiency and generalization \citep{brehmer2025does, xu2026efficient}. At the same time, hard constraints can become liabilities when the underlying symmetries are only approximate, as is common in real environments with boundaries, contact events, anisotropic friction, damping, and actuation. This tension motivates \emph{soft inductive biases}: parameterizations that nudge learning toward symmetry-respecting solutions while preserving capacity to fit structured violations of those symmetries~\citep{wilson2025deep,finzi2021residual}.

World models of objects moving in space are a natural substrate for such priors. By factoring scenes into entities and interactions, they promise compositional generalization and long-horizon stability~\citep{locatello2020object,wu2022slotformer}. Yet learning \emph{dynamics} (not just static properties) in multi-object scenes is brittle: contacts and constraints create sharp, state-dependent symmetry breaking, and naive, non-geometric architectures either overfit or require large data budgets. Conversely, strictly equivariant models can underfit when the data deviate from the assumed symmetries.

We argue that \emph{geometric algebra} (Clifford algebra)~\citep{roelfs2023graded} offers an effective middle ground for injecting geometric structure softly. By representing both states and operators as multivectors, projective geometric algebra (PGA) provides the model with a bias towards geometric transformations. The tensor nature of multivectors enables efficient implementation of standard modules such as linear maps, attention and nonlinearities~\citep{ruhe2023clifford, brehmer2023geometric}. Instead of enforcing exact equivariance end-to-end, our models use this soft geometric inductive bias to enable effective learning in environments with broken equivariance ~\citep{wilson2025deep, ruhe2023clifford}. The result is an \emph{object-centric dynamics model with soft geometric bias}: a next-step predictor trained autoregressively on object states and capable of generating long-horizon, physically plausible rollouts.

Practically, we embed each entity's state (position, velocity, and orientation) as multivectors in projective geometric algebras --- $\mathrm{PGA}(2)\cong Cl(2,0,1)$ for 2D scenes and $\mathrm{PGA}(3)\cong Cl(3,0,1)$ for 3D --- and train Clifford neural networks with soft geometric bias to predict sequences of these embedded object states. We evaluate our architectures as autoregressive dynamics models on procedurally generated 2D rigid-body scenarios from the JAX-based \textsc{Jax2D} engine~\citep{matthews2024kinetix}, 3D charged $N$-body simulations under harmonic confinement following~\citet{satorras2021n}, and real-world multi-agent driving trajectories from the Waymo Open Motion Dataset~\citep{ettinger2021large}. We show that the soft geometric bias improves both sample efficiency and long-horizon physical fidelity relative to (i) non-geometric MLP/Transformer baselines and (ii) a strictly equivariant variant of the Clifford transformer.

This perspective complements recent arguments that much of deep learning’s generalization can be understood through the lens of explicit priors and parameterization choices~\citep{wilson2025deep}, and it operationalizes ``soft equivariance'' ideas developed in residual pathway priors and related regularization schemes~\citep{finzi2021residual,kim2023regularizing}. It also connects to the emerging use of geometric algebra transformers across domains (e.g., $E(3)$ and Lorentz symmetry) which show that Clifford representations can unify objects and transformations within scalable attention architectures~\citep{brehmer2023geometric,spinner2024lorentz}.
\noindent Our main contributions are:

\begin{itemize}
    \item \textbf{Object-centric dynamics with soft geometric bias.} We introduce a Clifford-based dynamics model that biases features and interactions toward geometrically consistent behavior while allowing learned symmetry violations, yielding robustness in scenes with contacts, boundaries, and heterogeneous materials.
    \item \textbf{Autoregressive world modeling.} 
    We train a geometrically-informed transition model with a next-token prediction objective and evaluate long-horizon rollouts in multi-object scenes. The model is trained with a block-causal teacher forcing objective that has not been used in object-centric dynamics modeling before.
    \item \textbf{Empirical gains on synthetic and real-world data} On \textsc{Jax2D} rigid-body environments, 3D charged particles, and the Waymo Open Motion Dataset, our models achieve lower rollout errors and higher sample efficiency than non-equivariant baselines, and outperform or match a strictly equivariant Clifford counterpart when symmetries are only approximate.
\end{itemize}

\section{Related work}
\textbf{Soft inductive biases.} Existing work has explored the tension between hard inductive biases, such as strict equivariance, and softer parameterizations that accommodate broken symmetries in the data distribution. \citep{wilson2025deep, park2024approximate, huang2023approximately, petrache2023approximation, mcneela2023almost}. For example, Residual Pathway Priors convert hard constraints into priors that favor (but do not enforce) equivariance, improving robustness under misspecified symmetries~\citep{finzi2021residual}; related work regularizes toward mixed approximate symmetries~\citep{kim2023regularizing}, or specifically accommodates known violations of symmetries in otherwise equivariant models by modifying the input structure accordingly \citep{spinner2024lorentz}. \citet{wang2022approximately} learns to linearly combine a set of exactly equivariant kernels to relax equivariance constraints and identify broken symmetries. In contrast, we build our models using non-equivariant Clifford layers that can flexibly learn in varying degrees of symmetry.

\textbf{Equivariant architectures.} Existing equivariant architectures include SE(3)-Transformers and $E(n)$-GNNs, which enforce exact geometric equivariances in attention and message passing ~\citep{fuchs2020se,satorras2021n}. Geometric Algebra Transformers encode objects and operators in Clifford algebra with exact group equivariance in 3D or Lorentzian settings~\citep{brehmer2023geometric,spinner2024lorentz}. Here, we develop a variant of the geometric algebra transformer that also adopts GA primitives, but without enforcing strict equivariance, targeting settings where symmetry is intermittent or approximate. 

\textbf{Object-centric world models.} This work builds on existing literature on object-centric models, which decomposes high-dimensional sensory input (e.g., images or video) into exchangeable entities or objects, often referred to as `slots' \citep{burgess2019monet,greff2019multi, locatello2020object}. Further work extends this to the time domain by putting autoregressive dynamics onto the slot representations, which can be useful for identifying objects in the first place \citep{jaques2019physics} as well as for video prediction and planning ~\citep{kosiorek2018sequential,kipf2019contrastive,kossen2019structured,wu2022slotformer, ferraro2025focus, heins2025axiom}. In our setting, we assume we already have object-centric `slots' as inputs, but represent slot features in a projective geometric algebra and inject geometric constraints into their dynamics through Clifford neural networks. This is similar in spirit to other approaches which imbue slot dynamics with symmetries \citep{azari2022equivariant} or an explicit physics simulator \citep{jaques2019physics}. The seminal work introducing geometric algebra transformer (GATr) includes an 3-D \textit{n}-body prediction task, but is distinct in that their model is not a time-invariant transition model; rather, it is trained to predict final object states after a short period of integration from a stationary distribution of initial conditions \citep{brehmer2023geometric}. Additionally, the systems modeled in previous work \citep{brehmer2023geometric, satorras2021n} are truly equivariant systems, whereas in our work the presence of static, unrepresented objects like walls in the environment breaks strict equivariance when considering only the modeled, dynamic objects. An additional novelty of our work is the first use, to our knowledge, of block causal attention for object-centric modeling; previous work has used block-causal attention on encoded image patches as tokens, which do not explicitly represent objects \citep{liu2025bcat, dedieu2025improving, hafner2025training}.

\textbf{Benchmarks.} We use the \textsc{Jax2D} engine through the \textsc{Kinetix} framework to procedurally generate a set of rigid-body environments within constrained arenas, stressing models with approximate symmetries and contact dynamics~\citep{matthews2024kinetix}. To evaluate our models in 3D we use simulated environments of charged particles in a harmonic potential and constant gravitational acceleration \citep{satorras2021n}. Finally, to evaluate performance on real-world data, we target multi-agent driving scenarios through the Waymo Open Motion Dataset \citep{ettinger2021large}.

\section{Clifford transformer}
Clifford (geometric) algebras provide a unified algebraic framework in which geometric objects such as points, lines, and planes, as well as transformations such as rotations and translations, are represented as elements of the same algebra~\citep{roelfs2023graded}. This makes them a natural substrate for neural networks that should respect geometric structure: the geometric product encodes how objects compose under transformations, so that standard operations like linear maps and attention can be lifted to act on multivectors while retaining a geometric interpretation. Crucially, the same multivector primitives that support exactly equivariant operations also admit non-equivariant linear maps that retain a geometric parameterization without enforcing the symmetry --- this is the basis of the soft geometric inductive bias we develop here. We give a self-contained introduction to the 2D projective geometric algebra $Cl(2,0,1)$ in \cref{sec:geometric_algebra} and indicate how this construction carries over to the 3D case, including the construction of the soft geometric bias (\cref{sec:soft_geometric_bias}), and report architecture ablations in \cref{sec:model_ablations}.
Using these geometric algebra primitives, we implement a transformer architecture that acts on multiple entities per timestep, where each entity is represented by $C$ multivectors.

\begin{figure}[t]
    \begin{minipage}[t]{0.48\linewidth}
        \vspace{0pt}
        \centering
        \includegraphics[width=0.7\linewidth]{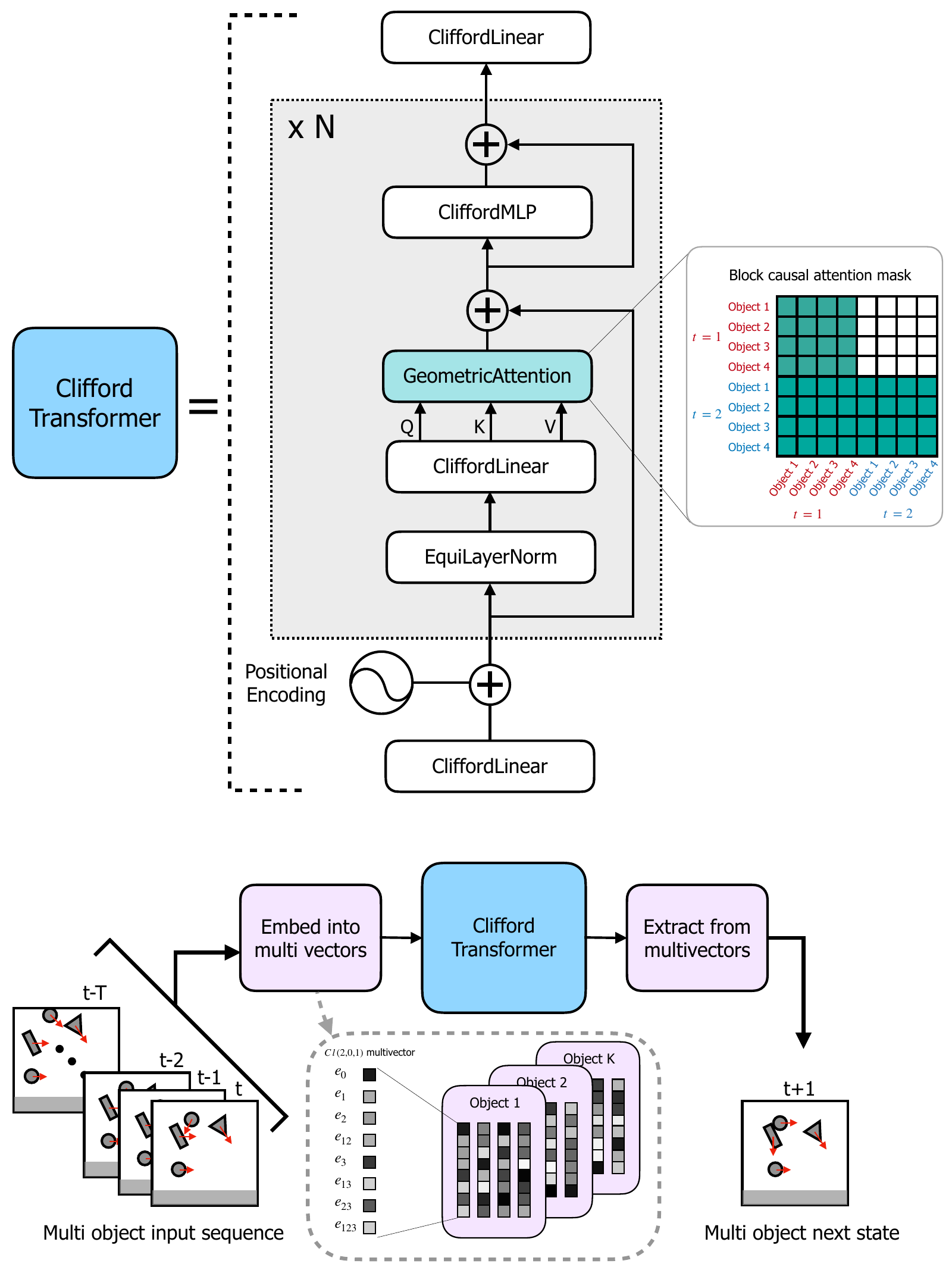}
        \caption{Top: Architecture diagram with the structure of $N$-block soft Clifford transformer models, including a block causal attention mask at the geometric attention layer. The block-causal attention mask is shown for an example of four objects and two timesteps, making a total of $4 \times 2 = 8$ input tokens. Bottom: the flow of information from multi-object representations to predicted next states.}
        \label{fig:architecture_diagram}
    \end{minipage}\hfill
    \begin{minipage}[t]{0.48\linewidth}
        \vspace{0pt}
        \centering
        \includegraphics[width=\linewidth]{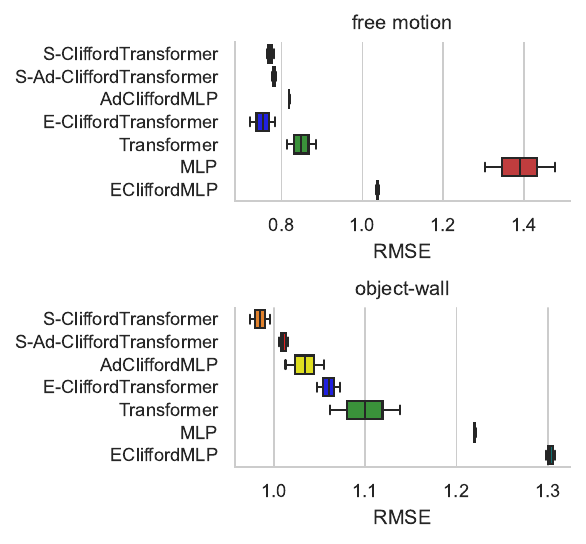}
        \caption{RMSE over 10 rollout frames against ground truth environment of 10 rigid-body polygons colliding in a box with gravity, separated by free motion (top) and object-wall collisions (bottom). Our Clifford models with soft geometric inductive bias (\texttt{\{S, S-Ad\}-CliffordTransformer}) outperform equivariant and baseline transformers on object-wall collisions, while matching on free motion.}
        \label{fig:rmse_per_type_10k_10block_non_equiv}
    \end{minipage}
\end{figure}

\subsection{Embedding}
In 2D, the position of the center of mass, as well as the position of polygon vertices, are embedded as multivectors using the $e_{13}$ and $e_{23}$ basis vectors. Object velocities are embedded using $e_{1}$ and $e_{2}$. Rotational angle, as well as angular velocity, is embedded using rotors $\cos(\theta) + \sin(\theta) e_{12}$. Each of these quantities is thus embedded as a separate multivector so that for example a circle with center of mass position $(x,y)$, velocity $(vx, vy)$, rotation angle $\theta$ and angular velocity $\dot{\theta}$, is represented as four multivectors $v_1 = x e_{13} + y e_{23}$, $v_2 = vx e_{1} + vy e_{2}$, $v_3 = \cos(\theta) + \sin(\theta) e_{12}$ and $v_4 = \cos(\dot{\theta}) + \sin(\dot{\theta}) e_{12}$.
This corresponds to one token in the input sequence where we view the 4 multivectors as 4 channels, each with 8 multivector components. A sequence of such tokens (for one object) is thus of the shape $(T, C, 8)$ where $T$ is the sequence dimension and $C$ is the multivector channel count. For $K$ different objects we have an input sequence of shape $(T, K, C, 8)$, which represents the sequence of each object's multivector tokens over time. In \cref{sec:arch} we describe how we deal with the $T$ and $K$ dimensions jointly during the attention operation.
This construction carries over in a straightforward manner to the 3D projective algebra $Cl(3,0,1)$, where the bivector rotations are now equivalent to the quaternion formulation.

\subsection{Clifford MLP}
\label{sec:clifford_mlp}
Using the building blocks of the geometric algebra, an extension of a standard linear layer can be constructed using multivector weights and features
\begin{equation}
    \mathrm{S}(x)^i = W^{ij} \star x_j,
\end{equation}
where $W$ is the learnable weight matrix with multivector coefficients, and $x=(x_1, x_2, ..., x_n)$ with $x_i\in Cl(2,0,1)$ is a collection of multivectors. For non-linear activation, we use a gated sigmoid $a(x) = \sigma(W^{\mathrm{act}}_i x^i)\, x$.

A Clifford multilayer perceptron (CMLP) can then be constructed with the notation $(n_1, n_2, ..., n_N)$ corresponding to $N$ linear layers followed by gated sigmoid activation after all but the last layer.  

We also evaluate models (AdCliffordMLP) with adjoint linear layers where the multivector weights acts through the adjoint sandwich product. To simplify the implementation we follow prior work~\citep{ruhe2023clifford} and use the Clifford reverse instead of the Clifford inverse.

\begin{equation}
    \mathrm{S}_{\mathrm{Ad}}(x)^i = W^{ij} \star x_j \star (W^{ij})^r
\end{equation}
where $(\cdot)^r$ is the Clifford reverse.

\subsection{Geometric algebra attention}
\label{sec:gattention}
Following \citet{brehmer2023geometric}, we use one of the natural induced inner products on $Cl(2,0,1)$ to implement multi-head attention. Explicitly, the inner product between two multivectors $x=x_0 + x_1 e_1 + \ldots + x_{123}e_{123}$ and $y=y_0 + y_1 e_1 + \ldots + y_{123}e_{123}$ takes the form
\begin{equation}
    x\cdot y = x_0 y_0 + x_1 y_1 + x_2 y_2 + x_{12} y_{12},
\end{equation}
with no contribution from the projective basis vector $e_3$.

Using three linear layers $Q, K, V$, self-attention is implemented with
\begin{equation}
    \mathrm{att}(x) = \frac{1}{\sqrt{d C}}\mathrm{softmax}_j\left(Q(x)_i\cdot K(x)_j\right) V(x)_j,
\end{equation}
where $Q, K, V$ output multivectors, d the dimension of the Clifford algebra (8 for Cl(2,0,1) and 16 for Cl(3,0,1)), and C are the number of multivectors in the embedding.

\subsection{Architecture}
\label{sec:arch}
We use three main variations of a transformer based on the Clifford primitives. The different versions use three different implementations of the MLP used throughout the model. The first uses a Clifford MLP as described in section \ref{sec:clifford_mlp}, referred to as \texttt{S-CliffordTransformer}, giving a soft geometric inductive bias. The second uses a linear layer where the multivector weights act on the input by the adjoint map, referred to as \texttt{S-Ad-CliffordTransformer}, giving a soft inductive bias with the ability to directly parametrize a geometric transformation.  The third is an exactly equivariant model using an equivariant linear map \citep{brehmer2023geometric} 
with rank-wise linear combinations, referred to as \texttt{E-CliffordTransformer}. 

All transformers are trained with teacher-forced next-token prediction over sequences of length $S$. The Clifford transformer starts with an embedding of the object centric variables as multivectors, followed by a channel expansion using a linear layer into shape $(B, S, K, C, 8)$, corresponding to batch ($B$), sequence position ($S$), object index ($K$), multivector channel ($C$) and the 8 dimensions of $Cl(2,0,1)$ (or in 3D, the 16 dimensions of $Cl(3,0,1)$). After channel expansion, we apply a sinusoidal positional embedding to the sequence position,
with frequency constant adapted to the sequence lengths relevant for the task. After this, the sequence and object dimensions are flattened into a single $N = S\times K$ dimension leaving a tensor of shape $(B, N, C, 8)$ in 2D, and $(B, N, C, 16)$ in 3D. This means that each token for the subsequent attention operation represents an object at a particular sequence position.

This is followed by a number of attention blocks, where each block performs causal multi-head self-attention (MHA).
\begin{align}
\begin{split}
    z &= \mathrm{CliffordMHA}(x) + x \\
    y &= \mathrm{CliffordMLP}(z) + z
\end{split}
\end{align}
The multi-head attention uses the Clifford algebra attention from section \ref{sec:gattention}, and a block-causal mask so that all objects at a given timestep are predicted in parallel rather than sequentially ~\citep{dedieu2025improving}. This also means that the model can reason jointly about all tokens within a given timestep when predicting the next timestep. In the case of generating rollouts in object-centric space, the Clifford transformers decode their outputs back into object-centric coordinates. When training these models, however, we compute and differentiate an $\mathrm{L}_2$ loss directly in the predicted multivector space, without decoding back to object-centric variables. The Waymo dataset features a  binary `active/inactive' flag for certain actor tokens, so we modify the $\mathrm{L}_2$ objective (used only for continuous variables) with a binary cross entropy loss for predicting the activity flag for the tokens that have it.

See Table \ref{tab:clifford_arch} for the transformer hyperparameters. For the baseline transformer we use exactly the same architecture, but with embedding dimension matched against total parameter count. See Appendix for further model ablations.
\begin{table}
    \centering
    \begin{tabular}{cccc}
    \toprule
         Blocks ($B$) & Heads ($H$) & MV ($C$) & $N_p$ \\
         \midrule
          10 & 8 & 24 & 1.5M\\ 
         \bottomrule
    \end{tabular}
    \caption{Clifford transformers hyperparameters with number of multivector channel $C$ and number of trainable parameters $N_p$.}
    \label{tab:clifford_arch}
\end{table}

\section{Datasets}
\textbf{Kinetix.}
We use the \textsc{Jax2D} engine ~\citep{matthews2024kinetix} to flexibly configure 2D physics environments with a desired number of objects and physical properties. 
All environments consist of dynamic rigid objects, either circles or rectangles, confined to a box and under the influence of gravity. The dynamic objects may collide with each other and with the static walls of the box. In all environments, the coefficients of friction and restitution are 1.0 for object-object contacts; the static walls use a restitution of 0.9 for stability. We find that the integration scheme of \textsc{Jax2D} can result in unstable simulation without the use of some damping. Although \textsc{Jax2D} environments contain also joints and thrusters, as well as possible actions, we only use rigid body circles and rectangles for our datasets.
We define a training dataset by sampling $N$ episodes with $128$ frames per episode. For our experiments we use $N=100$ and $N=1000$, resulting in 12,800 frames and 128,000 frames respectively. Each episode is reset with uniform random values for the different state variables of the objects.

\textbf{Waymo Open Motion Dataset.} To evaluate on real-world data we use the Waymo Open Motion Dataset (WOMD)~\citep{ettinger2021large}, which provides 9-second driving scenarios at 10\,Hz containing actor trajectories and static map geometry. Each scenario is tokenized into a fixed-width sequence: up to $K_{\mathrm{actor}}=32$ actor tokens (vehicles, pedestrians, cyclists within 80\,m), $K_{\mathrm{lane}}=64$ lane-centerline segments (resampled at 5\,m arc length within 100\,m) and $K_{\mathrm{misc}}=24$ miscellaneous map tokens (road edges, crosswalks, stop signs), giving 120 tokens per timestep. Actor states are embedded as multivectors in $Cl(2,0,1)$: position using the $e_{13}$ and $e_{23}$ basis vectors, velocity using $e_1$ and $e_2$, and heading as a rotor $\cos(\theta/2)+\sin(\theta/2)\,e_{12}$, following the same embedding scheme as the Kinetix environments. Lane and map tokens are embedded analogously using midpoints, tangent and normal directions. Beyond the geometric channels, each token carries 51 categorical
features (e.g. actor type, lane attributes, traffic-light state) embedded on
the scalar ($e_0$) blade of additional multivector channels. The prediction task is next-timestep regression of actor position deltas and heading rotor deltas. We train on 4k and 16k scenarios.

\textbf{Charged particles with harmonic confinement.} To evaluate our models in 3D we use a charged $N$-body simulation following~\citet{satorras2021n}: 5 particles with Coulomb interactions in 3D, integrated with a leapfrog scheme. The system is natively $E(3)$-equivariant. To introduce controlled symmetry breaking we add a harmonic confining potential along the $z$-axis, $F_z = -k_z z$, which breaks the full rotational symmetry while preserving $E(2)$ equivariance in the $xy$-plane, as well as gravity. By varying the confinement strength $k_z \in \{0, 0.5, 1, 3, 10\}$ we obtain a family of environments that interpolate between exact and strongly broken $E(3)$ symmetry, providing a controlled testbed for soft versus strict equivariance. We use 3000 training trajectories of 50 frames each. We embed positions, velocities, and charges as multivectors in $Cl(3,0,1)$.

\section{Results}
\label{sec:results}
We train all models using a batch size of 128, and optimize with AdamW using a learning rate of $5\cdot 10^{-4}$. For the transformers and CliffordMLP, we train with weight decay $10^{-7}$ and for the baseline MLP $3\cdot 10^{-6}$.
All models are trained using an $\mathrm{L}_2$ loss, where we find it beneficial to train the Clifford models using the $\mathrm{L}_2$ loss directly in multivector space. The Clifford transformers and vanilla transformers both converge in 20h for Kinetix 10k, and slightly quicker for WOMD and the 3d charged particle experiments on a single A100 gpu. See Appendix Figure \ref{fig:sample_efficiency} for sample efficiency to matched performance where the Clifford transformers outperform the baselines.
For an overview of the numerical results, see Figure~\ref{fig:first_page_combined}, where our \texttt{S-CliffordTransformer} consistently performs best in terms of rollout metrics across the three tasks.
\subsection{Baselines}
We evaluate against two baseline models without any geometric inductive bias: standard MLPs (\texttt{MLP}) and transformers (\texttt{Transformer}). The MLPs are two-layer ReLU activated with parameter-matched layer widths.
The baseline transformer follows the same architecture as the Clifford transformer using standard inner-product attention and block causal masking on an embedding space with dimensions adapted to match the overall parameter count of the Clifford transformer. To separate the effect of geometric inductive bias from the transformer architecture, we also include Clifford MLPs using the Clifford geometric product and multivector weights.

\subsection{Metrics}
To assess the quality of the model prediction over a rollout of multiple frames we use a number of metrics starting with the rollout RMSE.
\begin{equation}
\mathrm{RMSE}_{N} = \sqrt{\frac{1}{N\cdot N_{vars}}\sum_{t=0}^N\sum_{i\in\mathrm{vars}} \left(y^i_t - \tilde{y}^i_t\right)^2}
\end{equation}
where $N$ is the number of frames from the start of the sequence, $y^i_t$ is the model prediction for variable $i$ at time $t$, and $\tilde{y}^i_t$ is the target ground truth from the environment.
To separate the effect of accumulated errors from the models ability to correctly predict the forward integration of the physics, we also evaluate the deviation from one forward simulation of the state using the ground truth environment, giving a measure of the local dynamics error.

\begin{equation}
   \mathrm{RMSE}^{\mathrm{Euler}}_{N} = \sqrt{\frac{1}{N\cdot N_{vars}}\sum_{t=0}^N\sum_{i\in\mathrm{vars}} \left(y^i_t - \hat{y}^i_t\right)^2}
\end{equation}
where $y^i_t$ is the prediction at time $t$ of variable $i$, and $\hat{y}_t$ is one environment step from the predicted state $y_{t-1}$.
For the WOMD experiments we report the Average Displacement Error (ADE), the mean Euclidean distance between predicted and ground truth positions over the rollout horizon:
\begin{equation}
\mathrm{ADE}_{N} = \frac{1}{N}\sum_{t=1}^{N} \sqrt{\sum_{i\in\{x,y\}} \left(y^i_t - \tilde{y}^i_t\right)^2}
\end{equation}
where $y^i_t$ is the model prediction for coordinate $i$ at time $t$, and $\tilde{y}^i_t$ is the ground truth.

\subsection{Kinetix}
 In Figure \ref{fig:rmse_per_type_10k_10block_non_equiv}, our models with soft geometric inductive bias are evaluated against a baseline transformer using the same number of blocks and heads with parameter matched embedding dimensions, as well as an exactly equivariant Clifford transformer. In the bottom panel, validation rollout RMSE is calculated over 10 frames for all dynamic variables (position, velocity, angle and angular velocity) for frames with object-wall collisions. Since information about the geometry of the wall is not encoded in the input, this presents a non-equivariant quantity that the models need to learn. Our soft models (\texttt{\{S, S-Ad\}-CliffordTransformer, AdCliffordMLP}) outperform the baseline transformer and exactly equivariant Clifford transformer in terms of rollout RMSE for these frames. On the other hand, the free motion dynamics is still mostly equivariant and here the equivariant \texttt{E-CliffordModel} has the lowest RMSE, but our soft models still perform on-par. Figure~\ref{fig:rmse_object_object_10p_10k_seq1_10b} contains the 10 frame rollout RMSE for the remaining frame type of object-object interactions complementing the free motion and object-wall types in Figure~\ref{fig:rmse_per_type_10k_10block_non_equiv}. Our \texttt{\{S, S-Ad\}-CliffordTransformer} outperforms both the equivariant and the transformer baseline.
 An example of predicted frames from a rollout compared to ground truth simulation can be seen in Figure~\ref{fig:kinetic_10rectangles}.

\begin{figure}[t]
    \centering
    \begin{minipage}[t]{0.48\linewidth}
        \centering
        \includegraphics[width=\linewidth]{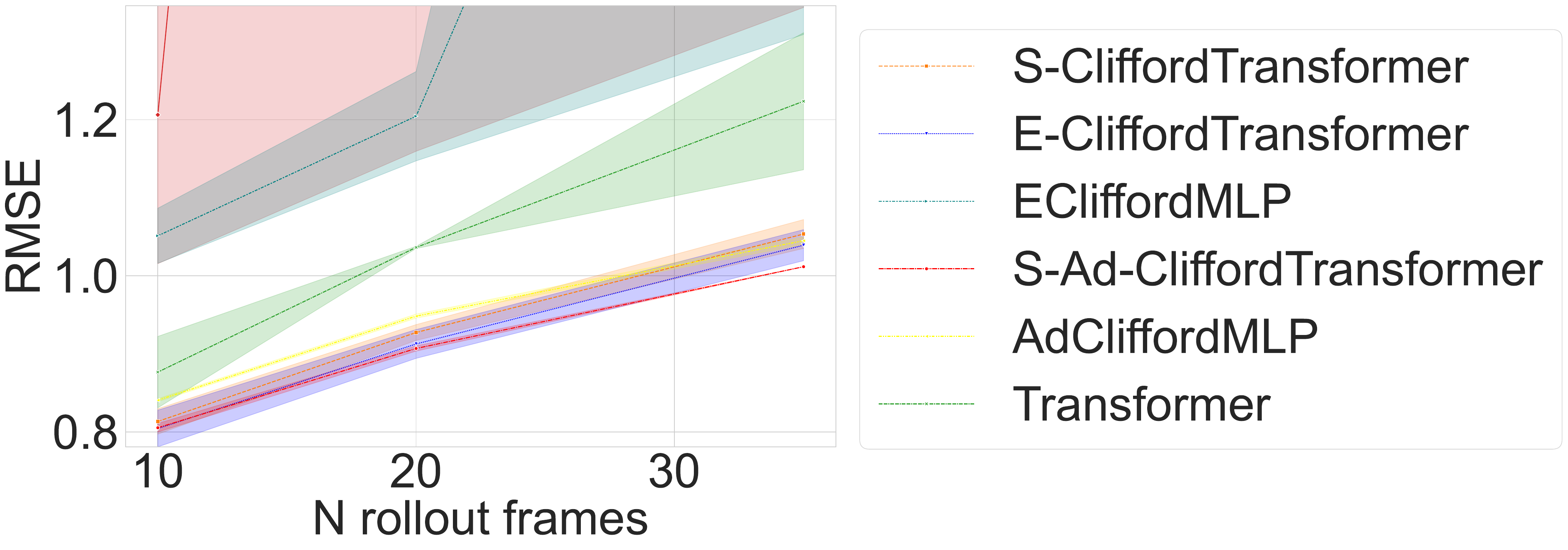}
        \subcaption{Long-horizon rollout RMSE for models trained on 10k frames with sequence length 1. All transformer models use the L architecture with 10-block, see section \ref{sec:arch}. Colored bands indicate 95\% CI intervals.}
        \label{fig:10k_10p_10b_seq1}
    \end{minipage}\hfill
    \begin{minipage}[t]{0.48\linewidth}
        \centering
        \includegraphics[width=\linewidth]{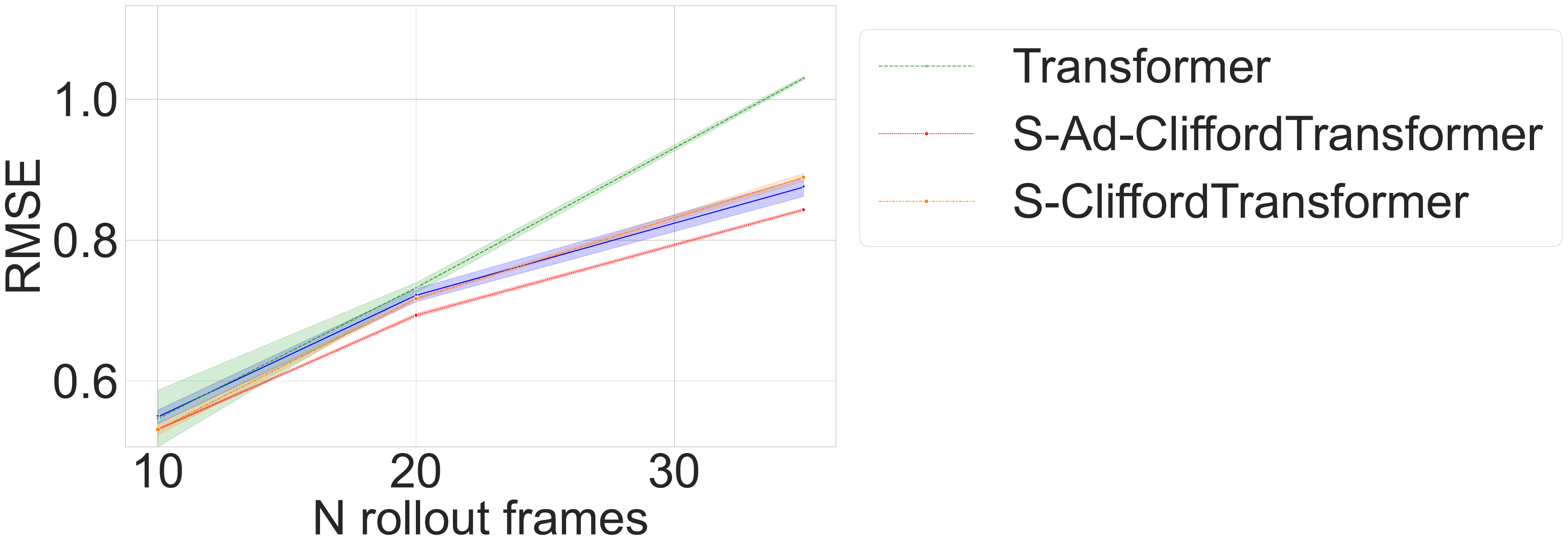}
        \subcaption{Rollout RMSE for models trained on 10k frames. All transformer models use a 10-block architecture with a causal sequence length of 2. With the additional context, the baseline transformer at this model scale can reach overall RMSE comparable to the our \texttt{\{S, S-Ad\}-CliffordTransformer} for shorter sequence lengths.}
        \label{fig:10k_10p_10b_seq2}
    \end{minipage}
    \label{fig:rollout_combined}
\end{figure}

We evaluate longer horizon rollout performance by RMSE of up to 35 frames. In Figure~\ref{fig:10k_10p_10b_seq1}, the rollout RMSE of our models compared to baselines are shown for sequence length 1, i.e. all models are performing next-frame predictions using only the dynamic variables from the current frame as input. The best performing model at 35 frames is our \texttt{S-Ad-CliffordTransformer}. The baseline MLP struggles to generalize, as well as the exactly equivariant Clifford MLP. The baseline transformer does learn some features of the dynamics, but there is a growing gap in total RMSE for longer horizons. In terms of the total RMSE, the equivariant transformer does well due to good performance on free motion frames; see the free motion panel in Figure~\ref{fig:rmse_per_type_10k_10block_non_equiv}. For longer horizons, \texttt{S-Ad-CliffordTransformer} surpasses the equivariant model also in terms of total RMSE.

Longer sequence context can be beneficial to dynamics models, and to evaluate how this affects performance we evaluate all transformer-based models first using 2 input frames with block causal attention. Figure~\ref{fig:10k_10p_10b_seq2} shows the rollout RMSE for all transformer models trained on 10k frames of 10 rigid body polygons. With the additional context of 2 frames, the baseline transformer matches performance of the soft and exactly equivariant models for shorter rollout lengths, but there is again a significant gap at longer rollouts. See appendix \ref{sec:additional_kinetix} for additional results on generalization, sequence length scaling, mixed object environments and sample efficiency.

\vspace{-0.3cm}
\subsection{Charged particles with harmonic confinement}
\begin{figure}[t]
    \centering
    \begin{minipage}[b]{0.28\linewidth}
        \centering
        \includegraphics[width=\linewidth, trim=0 50pt 0 30pt, clip]{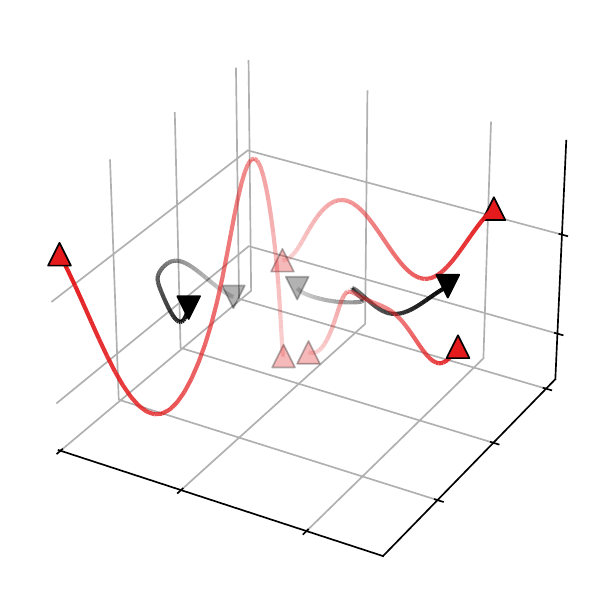}
        \caption{Example trajectories of 5 charged particles with harmonic confinement and gravity along $z$. Positive charges indicated by red traces and triangles pointing up, and negative as black pointing down.}
        \label{fig:nbody_trajectories}
    \end{minipage}\hfill
    \begin{minipage}[b]{0.68\linewidth}
        \centering
        \includegraphics[width=\linewidth]{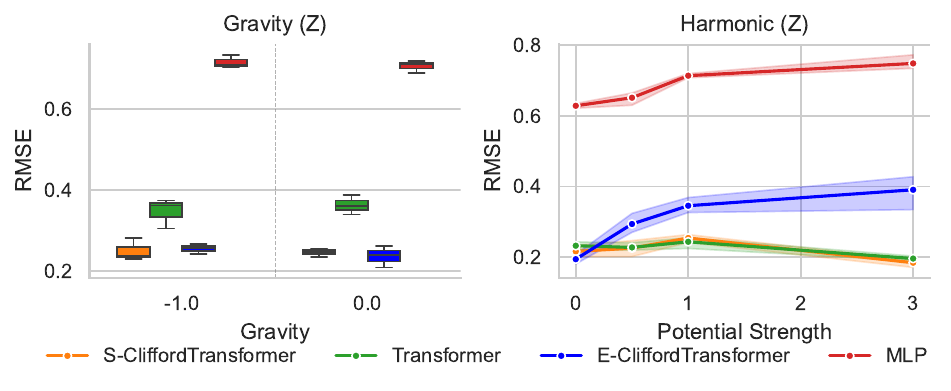}
        \caption{Symmetry-breaking ablations. Left: constant gravity along $z$. Right: harmonic confinement $F_z = -k_z z$ with increasing strength. The soft \texttt{S-CliffordTransformer} maintains robust performance as symmetry breaking increases, while the equivariant \texttt{E-CliffordTransformer} degrades under strong harmonic confinement.}
        \label{fig:nbody_ablations}
    \end{minipage}
    \label{fig:nbody_combined}
\end{figure}

Figure~\ref{fig:nbody_ablations} shows rollout RMSE for 3D charged particles under two symmetry-breaking ablations. 
For each particle, the 3D position and velocity are embedded as $x e_1 + y e_2 + z e_3 + e_0$ and $v_x e_1 + v_y e_2 + v_z e_3$ respectively, and the charge as a scalar multivector.
The left panel introduces constant gravitational acceleration along $z$, a uniform symmetry-breaking force: here the soft \texttt{S-CliffordTransformer} and the equivariant \texttt{E-CliffordTransformer} perform comparably. As the harmonic confinement increases in the right panel, the position-dependent restoring force progressively breaks $E(3)$ symmetry. The performance of the equivariant model degrades relative to \texttt{S-CliffordTransformer}, which maintains robust performance across all confinement strengths. The MLP baseline performs poorly throughout. Additional ablations on context length and prediction horizon are shown in Figure~\ref{fig:nbody_appendix} in the appendix.

\subsection{Waymo Open Motion Dataset}
\begin{figure}[t]
    \centering
    \begin{minipage}[b]{0.42\linewidth}
        \centering
        \includegraphics[width=0.9\linewidth]{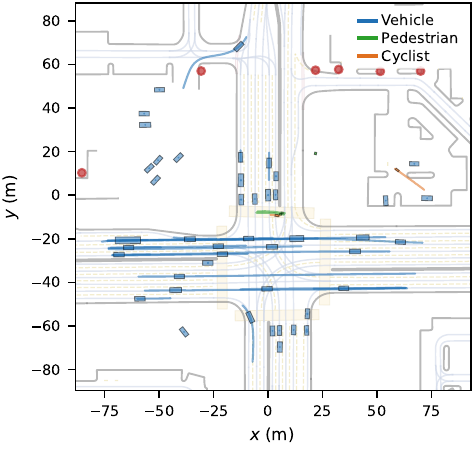}
        \caption{Bird's-eye view of an example WOMD scenario showing vehicle (blue), pedestrian (green), and cyclist (orange) trajectories at an intersection, with lane geometry and road boundaries.}
        \label{fig:womd_bev}
    \end{minipage}\hfill
    \begin{minipage}[b]{0.42\linewidth}
        \centering
        \includegraphics[width=\linewidth]{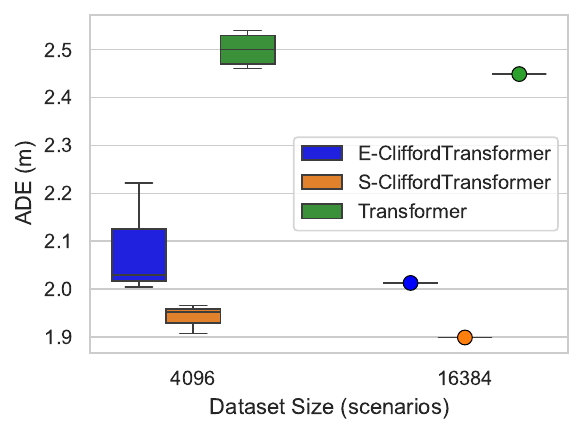}
        \caption{Average Displacement Error (ADE) at 1.6\,s on WOMD for 4k and 16k training scenarios. Boxes show the distribution over seeds (dots mark single-seed runs at 16k). The \texttt{S-CliffordTransformer} achieves the lowest ADE at both scales.}
        \label{fig:womd_rmse}
    \end{minipage}
\end{figure}

Figure~\ref{fig:womd_rmse} shows the position RMSE on the Waymo Open Motion Dataset as a function of dataset size. Both the soft (\texttt{S-}) and equivariant (\texttt{E-}) \texttt{CliffordTransformer} outperforms the baseline \texttt{Transformer} across dataset sizes, and the soft \texttt{CliffordTransformer} is consistently the best model. As the dataset grows to 16k scenarios all models improve, but the Clifford model maintains a clear advantage throughout, indicating that the geometric inductive bias provides a persistent benefit even on real-world driving data. Training curves are shown in Figure~\ref{fig:womd_training} in the appendix.

\vspace*{-0.3cm}
\section{Conclusion}
We introduced object-centric dynamics models with soft geometric bias through the use of geometric algebra primitives. Our soft Clifford transformers are simple to implement, and outperform exactly equivariant models as well as non-equivariant transformer baselines when trained to predict the dynamics of rigid body systems in 2D, charged $N$-body particles in 3D, and real-world multi-agent driving scenarios. In free‑motion regimes the strictly equivariant model remains competitive, but during symmetry‑breaking interactions our soft parameterizations maintain lower rollout error and reduced local dynamics error, indicating a better learned flow field.  Importantly, because we train these models autoregressively as next‑step predictors, the networks can roll out physically-realistic trajectories over long horizons without enforcing exact equivariance. This suggests the potential use of soft Clifford transformers as transition models in object-centric world models \citep{ferraro2025focus, kossen2019structured}. These results motivate the further use of soft geometric inductive bias as an effective tool to model real-world data distributions where symmetries are often approximate.

\vspace*{-0.3cm}
\section{Limitations}
\label{sec:limitations}
\vspace*{-0.2cm}
Although the inductive bias of using Clifford algebra neural networks is shown to perform well on a diverse set of benchmark datasets, there is still future work to be done in verifying their applicability on more data domains. If the task has no inherent geometric formulation, we do not expect this architecture to perform well. In this manuscript we have focused on autoregressive models, and we cannot claim that these results necessarily imply strong performance in a non-autoregressive setting. What is considered long horizon is relative to the complexity of task, when rolling out a next-step prediction model there is significant drift. Here we do not include e.g. noise~\cite{sanchez2020learning} during training so that we can clearly separate the effect of the geometric inductive bias. Our rollout horizons of e.g. 35 frames for Kinetix is long relative to the typical drift. We consider object-centric inputs and outputs to isolate the geometric aspects, making these models applicable for dense input and output maps is interesting future work. Regarding computational efficiency we have shown that the overall sample efficiency is better, as to be expected from the inductive bias, but this comes at the cost of a slight computational overhead in computing geometric tensor products. 
For both inference and training, it has been shown that similar architectures \citep{brehmer2025does,xu2026efficient} perform well in terms of performance under fixed FLOP budget.

\bibliographystyle{plainnat}
\bibliography{main}

\newpage
\appendix

\section{Additional Kinetix results}
\label{sec:additional_kinetix}

\begin{figure}[h]
    \centering
    \begin{tabular}{@{}c@{\hskip1pt}|@{\hskip1pt}c@{\hskip1pt}|@{\hskip1pt}c@{\hskip1pt}|@{\hskip1pt}c@{\hskip1pt}|@{\hskip1pt}c@{\hskip1pt}|@{\hskip1pt}c@{\hskip1pt}|@{\hskip1pt}c@{}}
    &$t{=}1$ & $3$ & $5$ & $7$ & $9$ & $11$ \\
    \hline
   \parbox[b]{0.4cm}{
  P\\[0.7cm]
  GT\\
 \vspace{0.1cm}
   } &\includegraphics[width=0.09\linewidth]{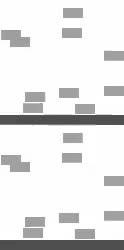} &
    \includegraphics[width=0.09\linewidth]{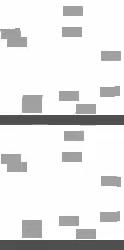} &
    \includegraphics[width=0.09\linewidth]{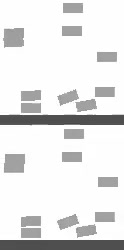} &
    \includegraphics[width=0.09\linewidth]{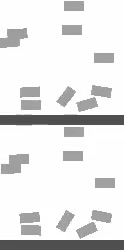} &
    \includegraphics[width=0.09\linewidth]{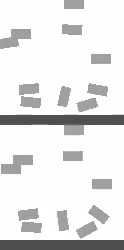} &
    \includegraphics[width=0.09\linewidth]{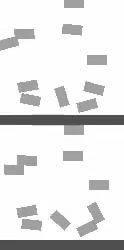}
    \end{tabular}
    \caption{Example rollout of ground truth (bottom) and model prediction (top) of frames 1 (left) to 11 (right) using our S-CliffordTransformer.
   The model predicts position, velocity, angle and angular velocity of the 10 rigid body polygons colliding with both the other dynamic bodies and the static environment with gravity.}
    \label{fig:kinetic_10rectangles}
\end{figure}

\begin{figure}[h]
    \centering
    \begin{minipage}[t]{0.48\linewidth}
        \centering
        \includegraphics[width=\linewidth]{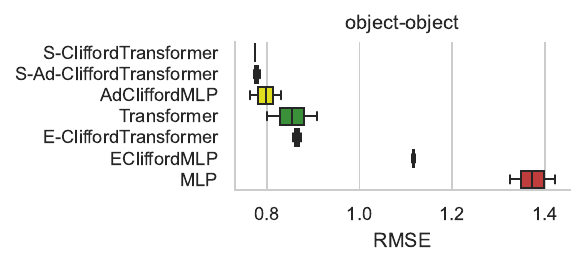}
        \caption{Rollout RMSE over 10 frames for object-object collision, complementing the two other modes illustrated in Figure~\ref{fig:rmse_per_type_10k_10block_non_equiv}. Models are ordered by their RMSE from best (\texttt{S-CliffordTransformer}) to worst (MLP).}
        \label{fig:rmse_object_object_10p_10k_seq1_10b}
    \end{minipage}\hfill
    \begin{minipage}[t]{0.48\linewidth}
        \centering
        \includegraphics[width=\linewidth]{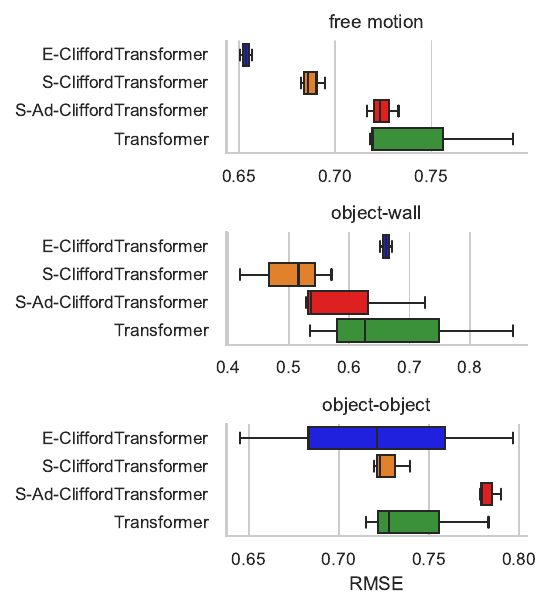}
        \caption{Rollout RMSE for 10 frames separated by frame type for all predicted variables. All models use input sequences of 2 frames and are trained using 10k simulated frames of 6 rigid body circles and 4 rigid body polygons.}
        \label{fig:10k_6b_4p_10b_seq2}
    \end{minipage}
\end{figure}

\begin{figure}[h]
    \centering
    \begin{minipage}[t]{0.32\linewidth}
        \vspace{0pt}
        \includegraphics[width=\linewidth]{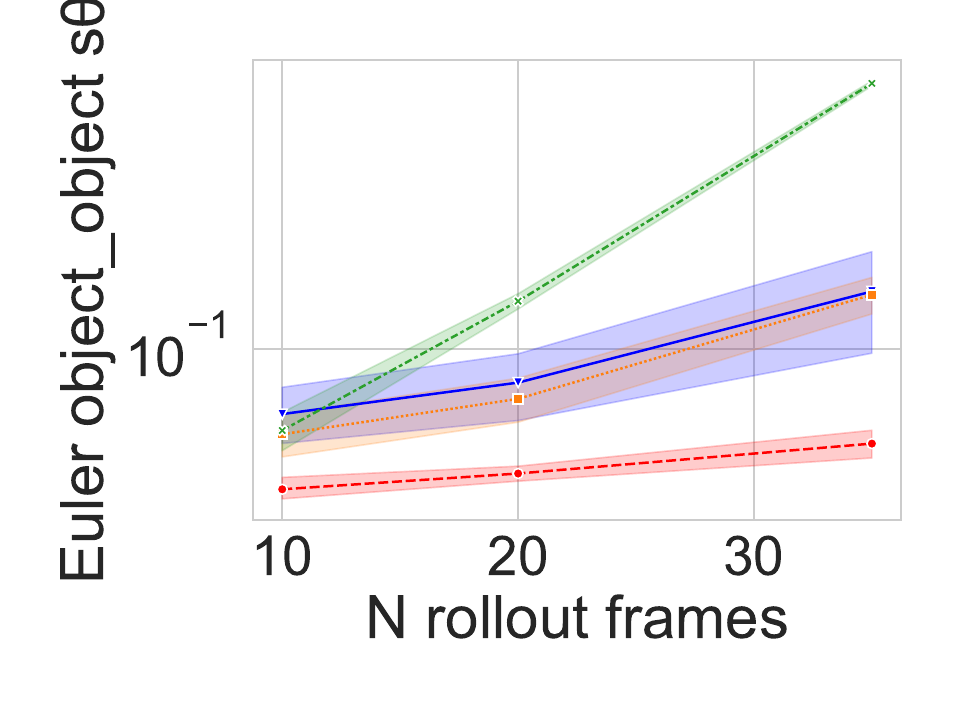}
        \includegraphics[width=\linewidth]{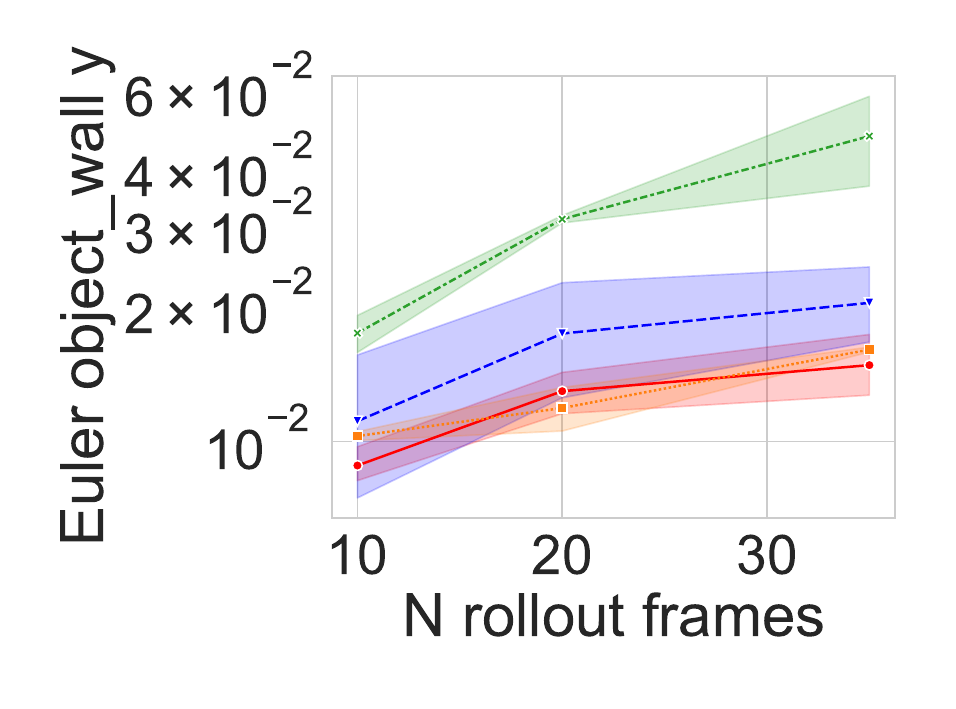}
        \subcaption{Euler RMSE for $\sin(\theta)$ on frames of object-object collisions (top) and $y$ on frames of object-wall interactions (bottom) for models trained on 10k frames. Colors indicate \texttt{Transformer} (green), \texttt{E-CliffordTransformer} (blue) and \texttt{\{S, S-Ad\}-CliffordTransformer} (orange and red).}
        \label{fig:euler_object_object_10k_b5_sine}
    \end{minipage}\hfill
    \begin{minipage}[t]{0.32\linewidth}
        \vspace{0pt}
        \includegraphics[width=\linewidth]{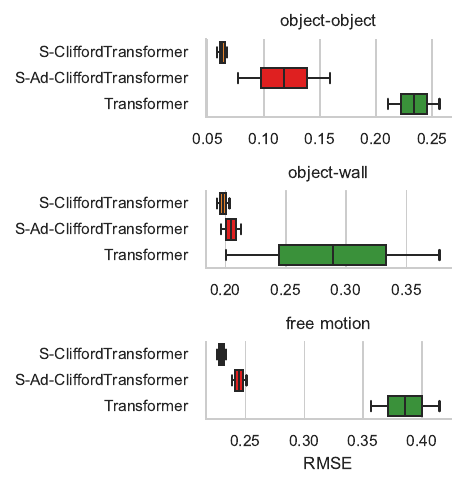}
        \subcaption{Rollout RMSE for transformer models with sequence length 16 and $N=1000$ episodes of 4 rigid body circles. Our models still outperform the baseline transformer in this regime, and attain significantly lower rollout error.}
        \label{fig:rmse_10b_16seq_1000ep}
    \end{minipage}\hfill
    \begin{minipage}[t]{0.32\linewidth}
        \vspace{0pt}
        \includegraphics[width=\linewidth]{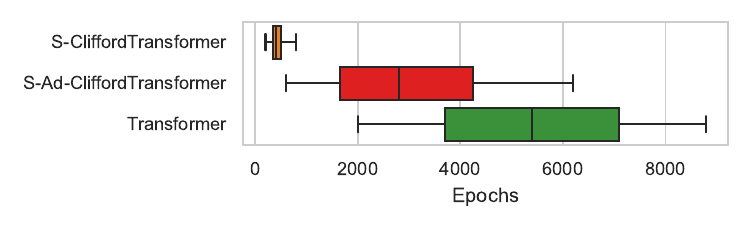}
        \subcaption{Epochs to match baseline transformer rollout RMSE at 10 frames. Models trained with sequence length 16 and $N=1000$ episodes.}
        \label{fig:sample_efficiency}
    \end{minipage}
\end{figure}
To further resolve how the soft inductive biases helps generalization, we evaluate the RMSE of next frame predictions against the environment ground truth forward step at each point along a long-horizon rollout. Figure \ref{fig:euler_object_object_10k_b5_sine} left panel shows the RMSE for one component $\sin(\theta)$ of the rotor prediction, i.e. the angular orientation of the polygons, on frames of object-object interactions. As object-object interactions are equivariant under $\mathrm{SE}(2)$ we expect the exactly equivariant models to perform well, which is reflected in the second best Euler RMSE after the soft Clifford transformer. For object-wall type interactions in Figure \ref{fig:euler_object_object_10k_b5_sine} right panel, the soft Clifford models perform better than the exactly equivariant model. This is again expected given the non-equivariance introduced by the static walls that are not part of the multi-object state provided to the models.
We also investigate the scaling behavior of the sequence length and dataset size in Figure~\ref{fig:rmse_10b_16seq_1000ep}, where our \texttt{\{S, S-Ad\}-CliffordTransformer} models still outperform the baseline transformer.

We also evaluate on environments with both polygons and discs. This means that the models have to learn non-trivial association of different moments of inertia and the effect of collision point on the resulting momentum and angular momentum. Figure \ref{fig:10k_6b_4p_10b_seq2} shows the total rollout RMSE over 20 frames for models trained on 10k frames using an input sequence length of 2. Note that this is twice as long as the rollout horizon in Figure  \ref{fig:rmse_per_type_10k_10block_non_equiv}. Between the Clifford based transformers, the trend observed for the 10 polygon case in Figure~\ref{fig:rmse_per_type_10k_10block_non_equiv} persists. For free motion frames, the equivariant \texttt{E-CliffordTransformer} has lower RMSE, whereas for interactions with other objects and static environment our \texttt{\{S, S-Ad\}-CliffordTransformer} performs better. At this longer rollout horizon the gap to the baseline transformer is increasing.

Finally, Figure~\ref{fig:sample_efficiency} shows that the soft inductive bias of the \texttt{S-CliffordTransformer} requires significantly fewer epochs to match the baseline transformer's rollout RMSE, confirming improved sample efficiency.

\section{Additional WOMD results}

Table~\ref{tab:womd_main} shows the next-step XY position RMSE on WOMD (4096 scenarios) broken down by agent type: SDC (self-driving car) and non-SDC (other agents). The \texttt{S-CliffordTransformer} achieves the lowest RMSE for both agent types. Tables~\ref{tab:womd_sdc} and~\ref{tab:womd_non_sdc} provide a per-variable breakdown. The soft Clifford model is best on position and velocity metrics, while the equivariant variant is competitive on heading prediction.

\begin{table}[h]
    \centering
    \caption{Next-step XY position RMSE on WOMD (4096 scenarios) by agent type. Bold indicates best.}
    \label{tab:womd_main}
    \begin{tabular}{llll}
\toprule
 & E-CliffordTransformer & S-CliffordTransformer & Transformer \\
Agent Type &  &  &  \\
\midrule
SDC & 0.0708 & \textbf{0.0334} & 0.1023 \\
Non-SDC & 4.6867 & \textbf{4.5321} & 4.8162 \\
\bottomrule
\end{tabular}

\end{table}

\begin{table}[h]
    \centering
    \caption{Per-variable RMSE breakdown for SDC agents on WOMD (4096 scenarios). Bold indicates best.}
    \label{tab:womd_sdc}
    \begin{tabular}{llll}
\toprule
 & E-CliffordTransformer & S-CliffordTransformer & Transformer \\
Variable &  &  &  \\
\midrule
$x$ & 0.0369 & \textbf{0.0163} & 0.0703 \\
$y$ & 0.0544 & \textbf{0.0254} & 0.0722 \\
$xy$ & 0.0708 & \textbf{0.0334} & 0.1023 \\
$v_x$ & 0.0314 & \textbf{0.0303} & 0.0664 \\
$v_y$ & \textbf{0.0651} & 0.0682 & 0.0775 \\
$v$ & \textbf{0.0777} & 0.0791 & 0.1028 \\
$\theta$ (rad) & 0.0084 & \textbf{0.0083} & 0.0099 \\
\bottomrule
\end{tabular}

\end{table}

\begin{table}[h]
    \centering
    \caption{Per-variable RMSE breakdown for non-SDC agents on WOMD (4096 scenarios). Bold indicates best.}
    \label{tab:womd_non_sdc}
    \begin{tabular}{llll}
\toprule
 & E-CliffordTransformer & S-CliffordTransformer & Transformer \\
Variable &  &  &  \\
\midrule
$x$ & 3.2378 & \textbf{3.1310} & 3.4162 \\
$y$ & 3.1245 & \textbf{3.0238} & 3.3811 \\
$xy$ & 4.6867 & \textbf{4.5321} & 4.8162 \\
$v_x$ & 0.4371 & \textbf{0.4326} & 0.7522 \\
$v_y$ & 0.4106 & \textbf{0.4010} & 0.7788 \\
$v$ & 0.6603 & \textbf{0.6514} & 1.0993 \\
$\theta$ (rad) & 0.2270 & 0.2337 & \textbf{0.2261} \\
\bottomrule
\end{tabular}

\end{table}

\begin{figure}[h]
    \centering
    \includegraphics[width=0.6\linewidth]{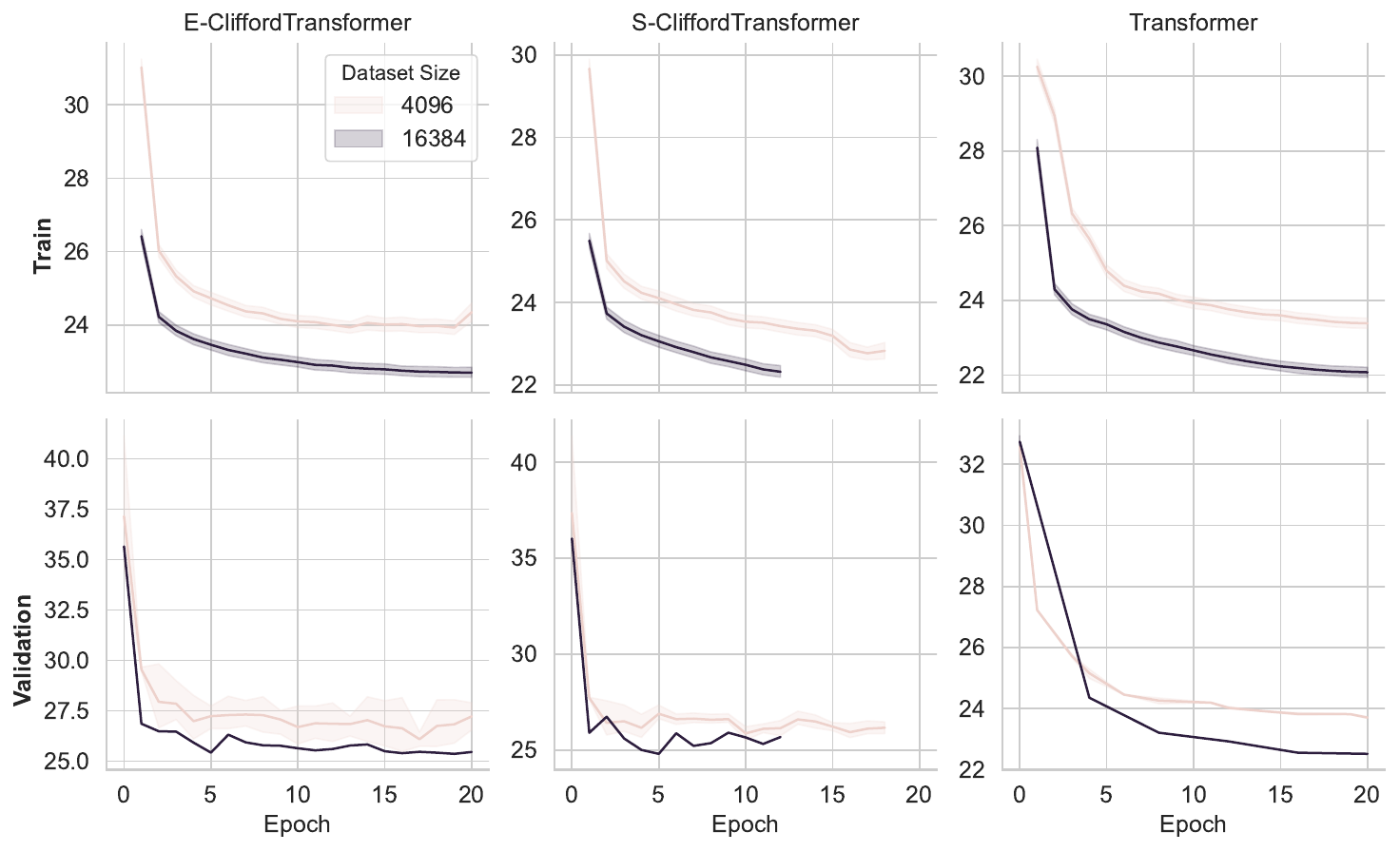}
    \caption{Training and validation loss curves on the Waymo Open Motion Dataset for the \texttt{E-CliffordTransformer} (left), \texttt{S-CliffordTransformer} (middle) and \texttt{Transformer} (right) across dataset sizes (4096 and 16384 scenarios).}
    \label{fig:womd_training}
\end{figure}

Figure~\ref{fig:womd_training} shows the training and validation loss curves for the WOMD experiments.

\section{Additional charged particle results}

\begin{figure}[h]
    \centering
    \includegraphics[width=0.7\linewidth]{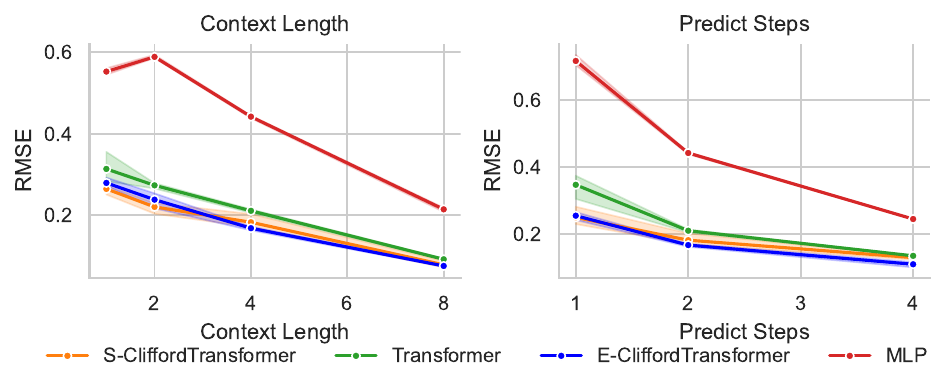}
    \caption{Additional ablations on the 3D charged particle system. Left: varying context length (number of input frames). Right: varying prediction horizon (number of steps predicted). The \texttt{S-CliffordTransformer} maintains robust performance across all settings.}
    \label{fig:nbody_appendix}
\end{figure}

Figure~\ref{fig:nbody_appendix} shows additional ablations on the 3D charged particle system beyond the gravity and harmonic confinement sweeps in the main text.

\section{Geometric Algebra}
\label{sec:geometric_algebra}
The projective geometric algebra $Cl(2,0,1)$ is spanned by 8 basis vectors $\left\{e_0, e_1, e_2, e_3, e_{12}, e_{13}, e_{23}, e_{123}\right\}$, often referred to as blades to disambiguate from the vector part of the multivectors ~\cite{roelfs2023graded}. The algebra is equipped with the geometric product, a bilinear product such that $e_i \star e_i = 1$ for $i\in \{1,2\}$ and $e_3 \star e_3 = 0$ for the projective dimension. Together with the anti-commutation relation $e_i \star e_j = - e_j \star e_i$ for $i\neq j$, this completely specifies the structure constants of the algebra. When writing the geometric product between basis blades we will omit the explicit $\star$, in fact the higher order basis blades are themselves just shorthand for the geometric product of the rank 1 basis blades $e_1$, $e_2$ and $e_3$.

The algebra acts on itself through the twisted adjoint action $g(x) = g^\star \star x \star g^{-1}$, with the grade involution $g^\star$ defined on monomials by $(e_1\ldots e_k)^\star = (-1)^k e_1 \ldots e_k$ and extended by linearity to polynomials. Through the twisted adjoint action each basis vector $e_i$ can be identified with reflections in an orthogonal plane. Using the Cartan–Dieudonné theorem this can be extended to all the roto-translations of the 2-dimensional Euclidean group $E(2)$.

Geometric entities like points, vectors and lines can be naturally embedded in the Clifford algebra, and their transformations under $E(2)$ by the twisted adjoint action. For example points can be represented by the grade 1 basis vectors, and vectors using the grade 2 basis vectors. A rotation by an angle $\theta$ is parametrized by the rotor $R=\cos(\theta/2) + \sin(\theta/2) e_{12}$ that acts on multivectors by $R^\star \star (x e_1 + y e_2) \star R^{-1}$, resulting in the rotated coordinates $(\cos(\theta)x - \sin(\theta)y)e_1 + (\sin(\theta) x + \cos(\theta) y)e_2$.
In terms of numerical tensors, a multivector in $Cl(2,0,1)$ will thus be an 8-dimensional real vector.

The geometric product between two multivectors can be calculated using the structure coefficients, i.e. the geometric product between the basis vectors, and using the linearity of the product.
\begin{align}
    x\star y &= (x^i e_i)\star(y^j e_j) \\
        &= x^i y^j (e_i \star e_j) \\
        &= x^i y^j f_{ij}^{k}e_k,
\end{align}
where the indices $i$ and $j$ range over the set of 8 multivector dimensions $\{0, 1, 2, 3, 12, 13, 23, 123\}$ of $Cl(2,0,1)$, and $f_{ij}^{k}$ are the structure constants 
\begin{equation}
   f_{ij}^k = (e_i \star e_j)^k 
   \label{eq:structure}
\end{equation}.

\subsection{Implementation}
The structure constants $f_{ij}^k$ of equation \eqref{eq:structure} can be precalculated for $Cl(2, 0, 1)$ as a tensor $f$ of shape $(8,8,8)$. Two multivectors $x$ and $y$, represented by their 8 components, can then be multiplied with
\begin{verbatim}
def clifford_product(x, y, f):
    return einsum("...i,...j,ijk->...k", 
                  x, y, f)
\end{verbatim}
where $f$ is the precomputed tensor containing the $8\times8\times8$ structure constants.

\subsection{Soft geometric bias}
\label{sec:soft_geometric_bias}
Through the use Clifford algebra primitives, our models are constructed to have a soft geometric inductive bias without enforcing strict equivariance. Other works, such as the Geometric Algebra Transformer (GATr) \cite{brehmer2023geometric} target the equivariant domain by using only equivariant operations. Although the geometric algebra provides a natural framework for equivariant models, it also provides a succinct parametrization of non-equivariant geometric transformations. 
To see why this is, recall that given an input $x\in X$ in a space with an action of a group $G$, a function $f: X \rightarrow Y$ is said to be equivariant if $f(g x) = g f(x)$, where $g\in G$ acts through a possibly different representation on the output space $Y$. In other words, if the input transforms, the output should follow along.

Since the Clifford inner product is by definition invariant under the action of $E(2)$, the attention block transforms in the same representation as the value map $V(x)$. In this work, we evaluate three different ways of parameterizing the MLP that is used for all projections in the models: the exactly equivariant linear map of GATr using rank-wise linear combinations, a Clifford linear map $S(x)^i = w^{ij} \star x_j$ and a Clifford adjoint linear map $S_{\mathrm{Ad}}(x)^i = w^{ij} \star x \star (w^{ij})^r$, where $(\cdot)^r$ denotes the reverse operation. Neither of these latter two linear maps are equivariant since the action of $E(2)$ on a multivector $x$ parametrized by another multivector $m$ is given by $m^r \star x \star m^{-1}$ and
\begin{align}
    S(m^r\star x \star m^{-1})^i &= w^{i j} \star m^r\star x_j \star m^{-1} \\
    &\neq m^r \star w^{i j} \star x_j \star m^{-1} \\
    &= m^r \star S(x) \star m^{-1},
\end{align}
by the non-commutativity of the geometric product. An almost identical computation shows this to be the case for $S_{\mathrm{Ad}}$ as well. As such, our models using the linear maps $S$ and $S_{\textrm{Ad}}$ have a soft geometric inductive bias without enforcing strict equivariance.

\section{Model ablations}
\label{sec:model_ablations}
\begin{figure}
    \centering
    \includegraphics[width=1.0\linewidth]{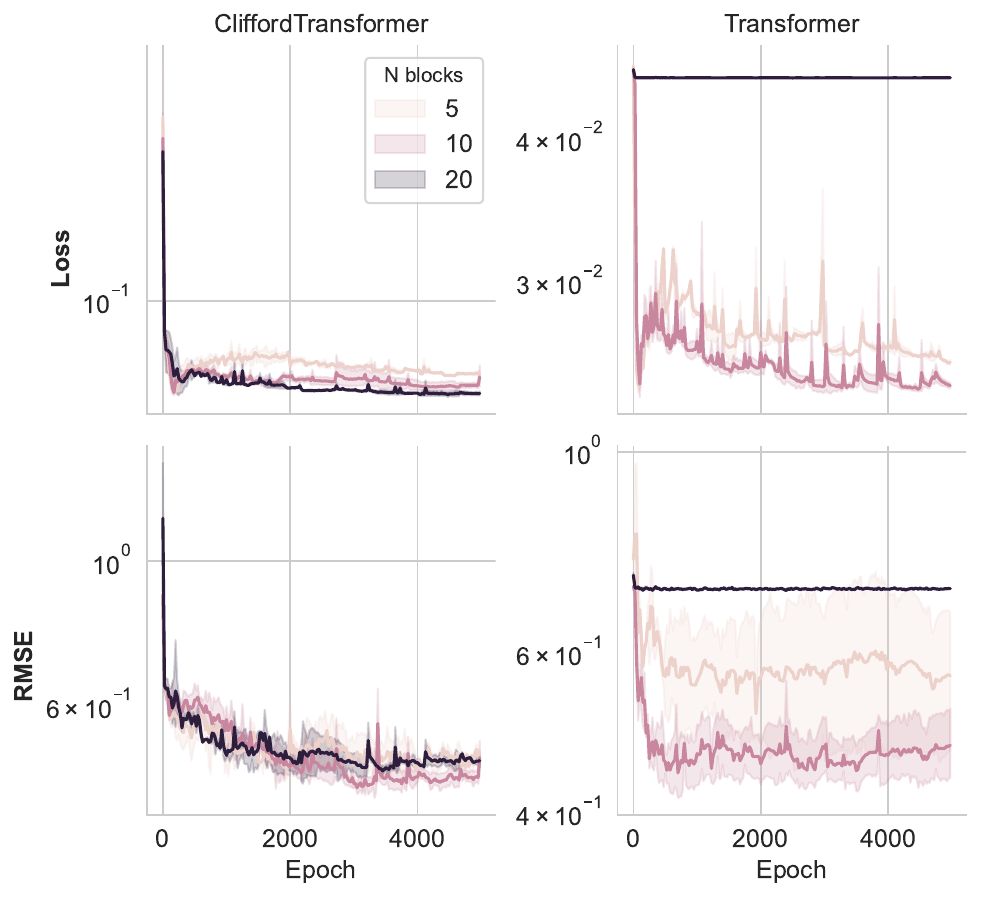}
    \caption{Validation loss and 10 frame rollout RMSE during training for different number of transformer blocks (5, 10, 20) for our \texttt{\{S, S-Ad\}-CliffordTransformer} models and baseline \texttt{Transformer}. Trained on 100 episodes with sequence length of 16.}
    \label{fig:ablate_blocks}
\end{figure}

\begin{figure}
    \centering
    \includegraphics[width=1.0\linewidth]{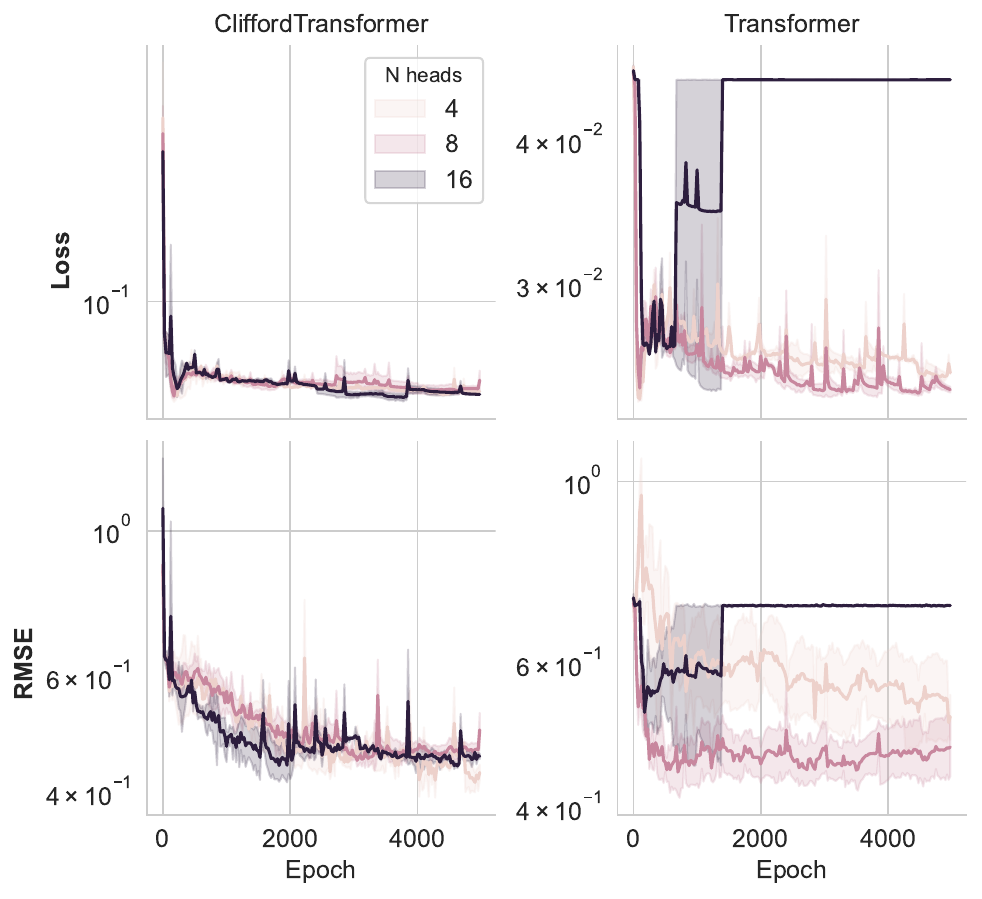}
    \caption{Validation loss and 10 frame rollout RMSE during training for different number of transformer heads (4, 8, 16) for our \texttt{\{S, S-Ad\}-CliffordTransformer} models and baseline \texttt{Transformer}. Trained on 100 episodes with sequence length of 16.}
    \label{fig:ablate_heads}
\end{figure}

To select the main architecture hyperparameters (number of blocks, number of heads, number of internal channels) we performed a number of ablations for both our \texttt{\{S, S-Ad\}-CliffordTransformer} models and for the baseline \texttt{Transformer}. After an initial exploration phase we identified favorable regions for these three hyperparameters. First, for number of blocks in Figure~\ref{fig:ablate_blocks}, both our models and the baseline transformer are stable under our training schedule up to 10 blocks, and our \texttt{CliffordTransformer} models continue to show stable training at 20 blocks. In terms of rollout RMSE, the baseline transformer performs best at 10 blocks which is the value chosen for all the main experiments. Second, for number of transformer heads in Figure~\ref{fig:ablate_heads} we again find a similar pattern, with the baseline transformer having a preferred value of 8 heads among the ablated values. Again, we choose this favorable point for the baseline transformer as the value used for all the main experiments. Third, for the internal channel counts we do a separate ablation starting from a parameter matched point. Since the Clifford-based transformers have more parameters per channel on account of the 8 multivector dimensions, we cannot simply match the overall channel count. Similarly to the other main hyperparameters, we choose 64 channels as a point where the baseline transformer performs well here and then we parameter match our \texttt{CliffordTransformer} models at this point resulting in 24 multivector channels. In summary, the overall architecture hyperparameters resulting from this ablation for our \texttt{CliffordTransformer} models are shown in Table \ref{tab:clifford_arch} in the main manuscript.

\begin{figure}
    \centering
    \includegraphics[width=0.5\linewidth]{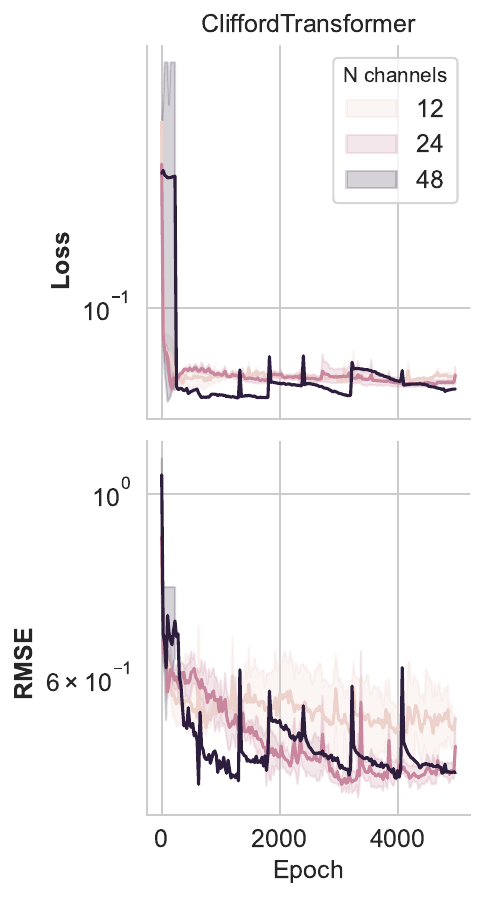}
    \caption{Validation loss and 10 frame rollout RMSE during training for different number of multivector channels (12, 24, 48) for our \texttt{\{S, S-Ad\}-CliffordTransformer} models. Trained on 100 episodes with sequence length of 16.}
    \label{fig:ablate_clifford_channels}
\end{figure}

\begin{figure}
    \centering
    \includegraphics[width=0.5\linewidth]{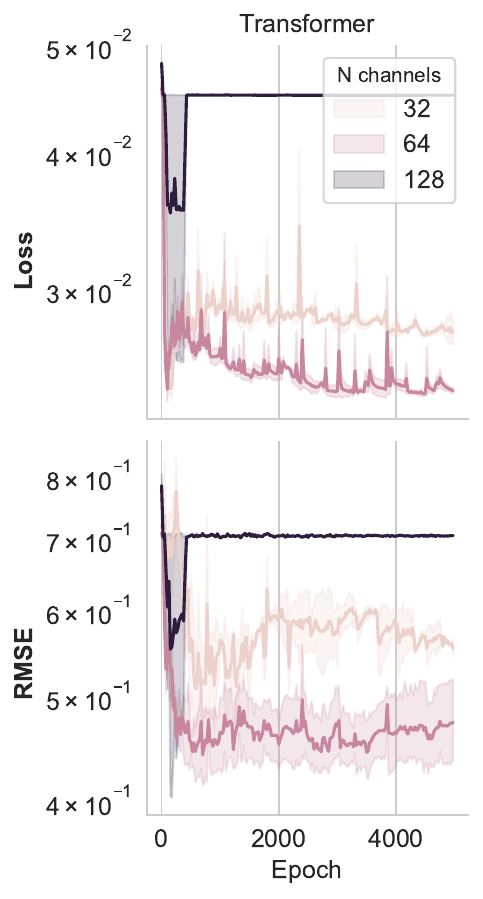}
    \caption{Validation loss and 10 frame rollout RMSE during training for different embedding dimensions (32, 64, 128) for the baseline \texttt{Transformer}. Trained on 100 episodes with sequence length of 16.}
    \label{fig:ablate_vanilla_channels}
\end{figure}

\section{Qualitative results}
We also include a number of qualitative results in the form of rollouts from an initial environment state for environments with 10 circles, 4 circles, 2 circles and 2 rectangles and 10 rectangles. The rollouts are chosen to highlight the typical behavior of the models over many rollouts.

\begin{figure}
    \centering
    \includegraphics[width=0.8\linewidth]{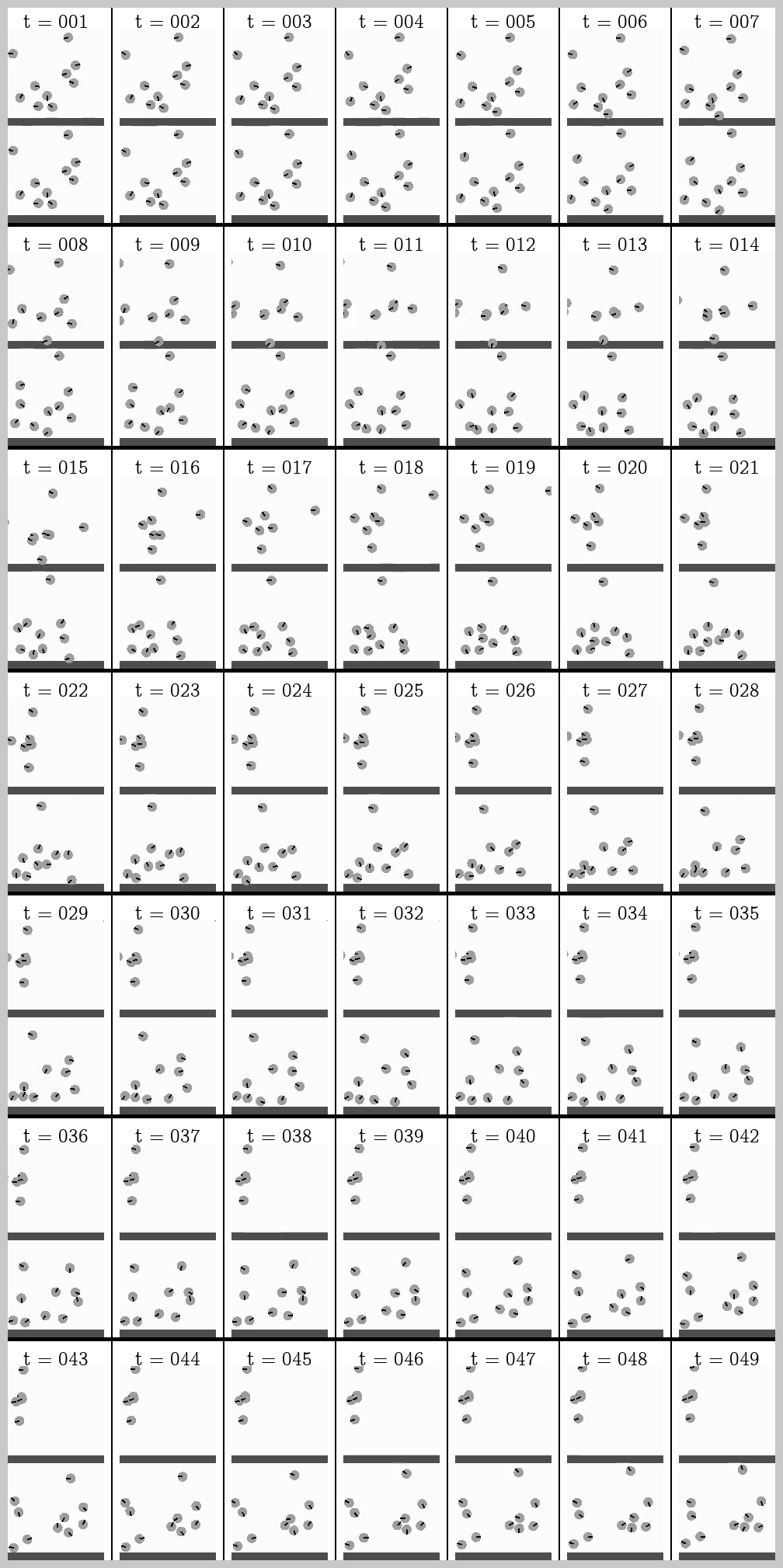}
    \caption{Example rollout using baseline transformer trained on 1000 episodes of 10 circles using sequence length 16. For each timestep, model prediction is shown in the top panel, and the ground truth environment simulation in the bottom panel.}
    \label{fig:supp_rollout_transformer_10balls_16seq}
\end{figure}

\begin{figure}
    \centering
    \includegraphics[width=0.8\linewidth]{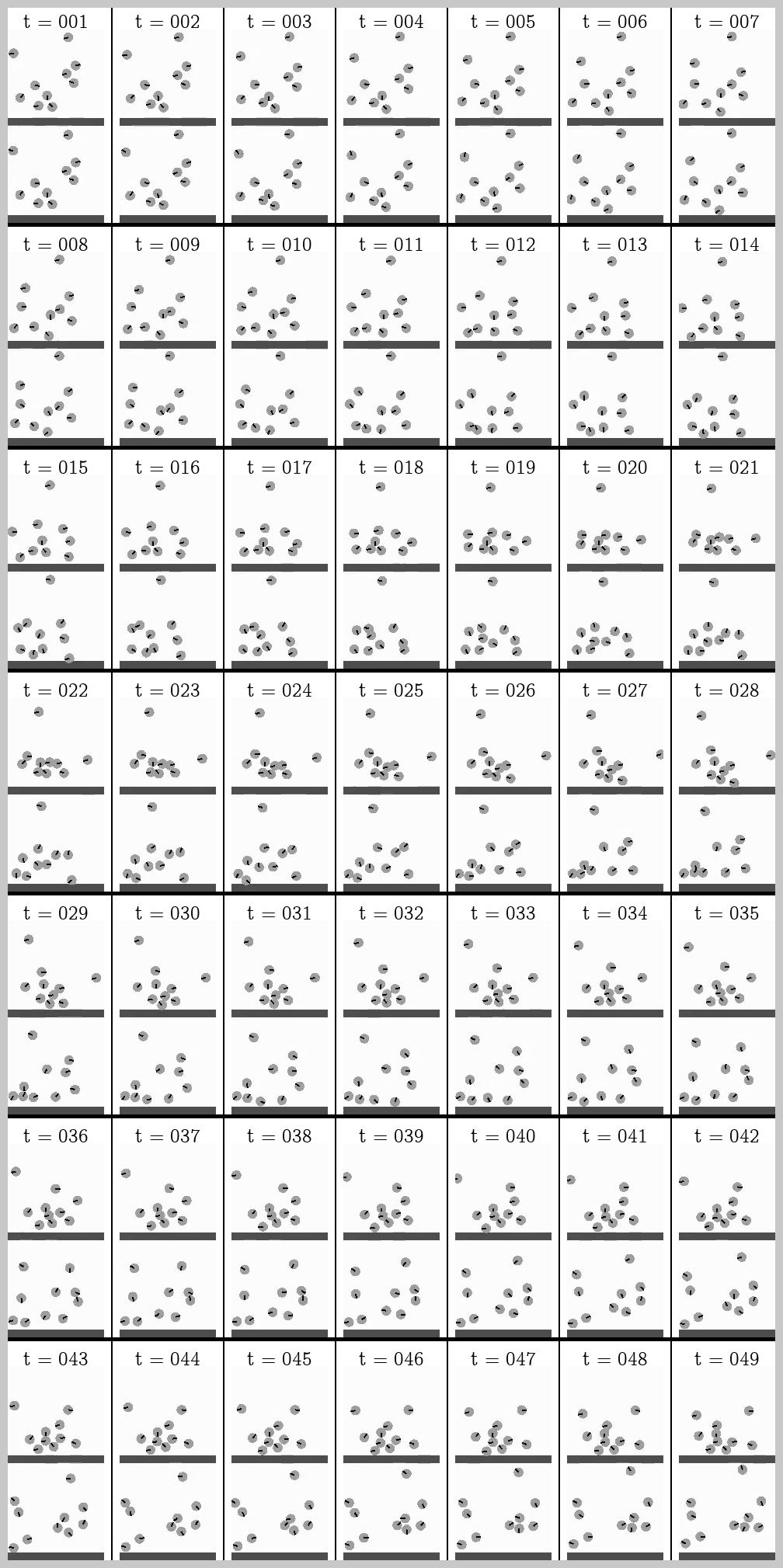}
    \caption{Example rollout using \texttt{S-Ad-CliffordTransformer}, using adjoint linear layers, trained on 1000 episodes of 10 circles using sequence length 16. For each timestep, model prediction is shown in the top panel, and the ground truth environment simulation in the bottom panel.}
    \label{fig:supp_rollout_ad_cliffordtransformer_10balls_16seq}
\end{figure}

Starting with the 10 circles environment, Figure \ref{fig:supp_rollout_transformer_10balls_16seq} shows a rollout using the baseline \texttt{Transformer}, and Figure \ref{fig:supp_rollout_ad_cliffordtransformer_10balls_16seq} using our \texttt{S-Ad-CliffordTransformer}. Note that the baseline predictions results in static object penetration, e.g. starting at timestep 7. Furthermore, while our Clifford transformer in Figure \ref{fig:supp_rollout_ad_cliffordtransformer_10balls_16seq} does a robust rollout with good qualitative agreement up to timestep 20, and rough agreement all the way up to timestep 49, the baseline transformer quickly diverges.

\begin{figure}
    \centering
    \includegraphics[width=0.8\linewidth]{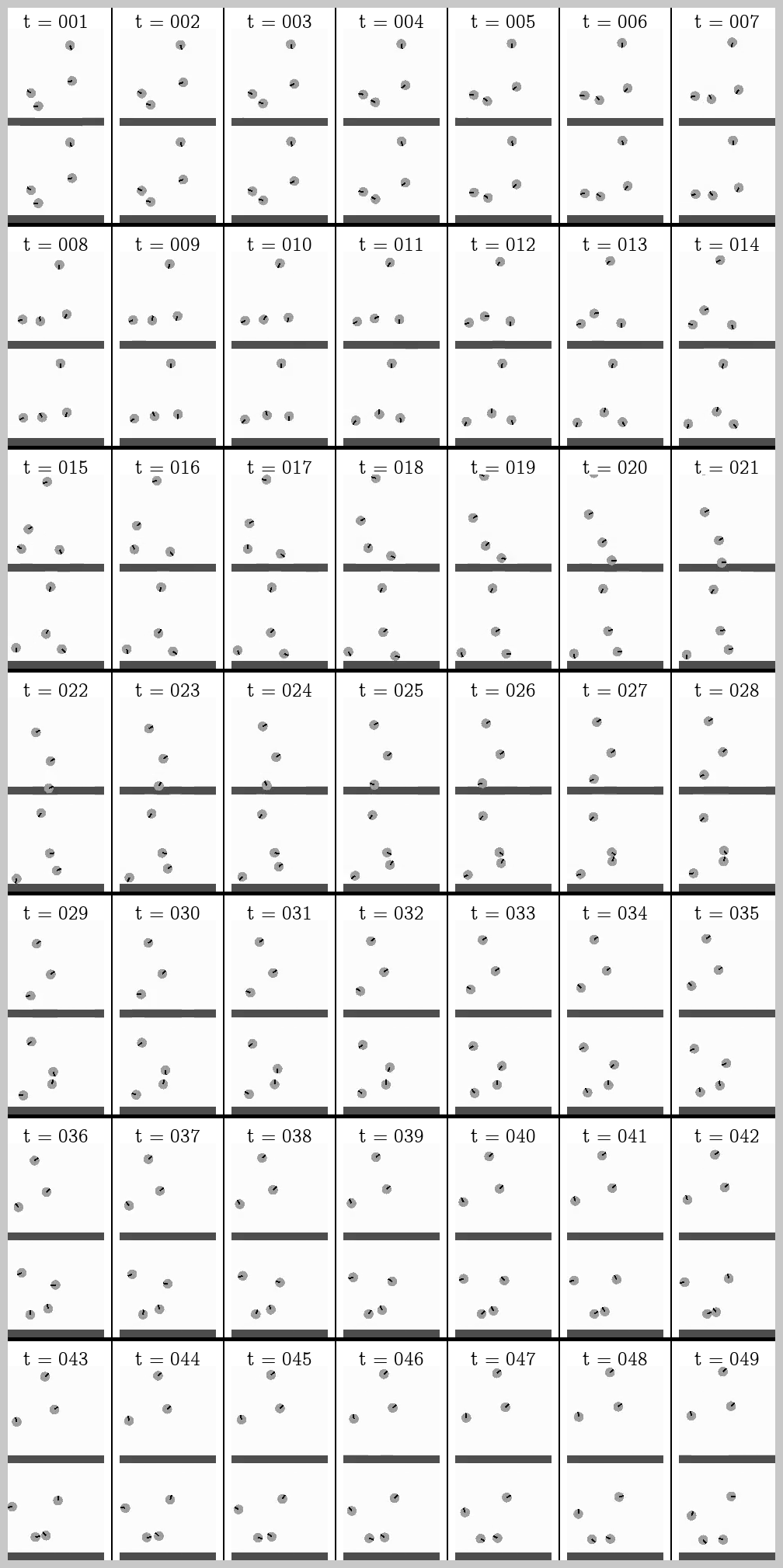}
    \caption{Example rollout using baseline transformer trained on 1000 episodes of 4 circles using sequence length 16. For each timestep, model prediction is shown in the top panel, and the ground truth environment simulation in the bottom panel.}
    \label{fig:supp_rollout_transformer_4balls_16seq}
\end{figure}

\begin{figure}
    \centering
    \includegraphics[width=0.8\linewidth]{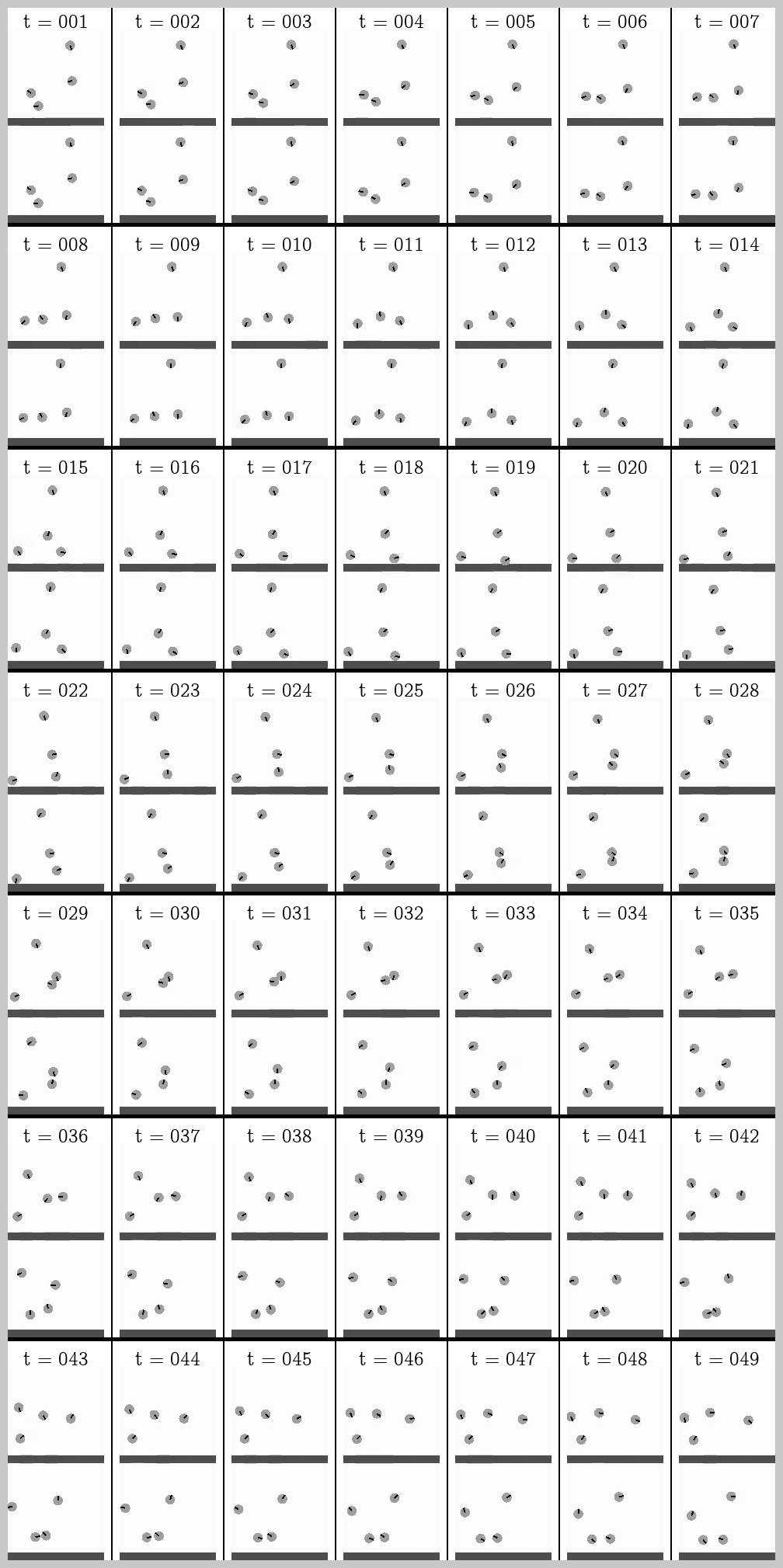}
    \caption{Example rollout using \texttt{S-Ad-CliffordTransformer}, using adjoint linear layers, trained on 1000 episodes of 4 circles using sequence length 16. For each timestep, model prediction is shown in the top panel, and the ground truth environment simulation in the bottom panel.}
    \label{fig:supp_rollout_ad_cliffordtransformer_4balls_16seq}
\end{figure}

Continuing with 4 circles, our Clifford transformer in Figure \ref{fig:supp_rollout_ad_cliffordtransformer_4balls_16seq} captures the dynamics almost perfectly up to timestep 28 where a collision occurs. The collision is detected and resolved, although slightly different from the ground truth environment. In contrast, the baseline transformer  in Figure \ref{fig:supp_rollout_transformer_4balls_16seq} fails to predict the free motion dynamics such that the collision occurs at all.

\begin{figure}
    \centering
    \includegraphics[width=0.8\linewidth]{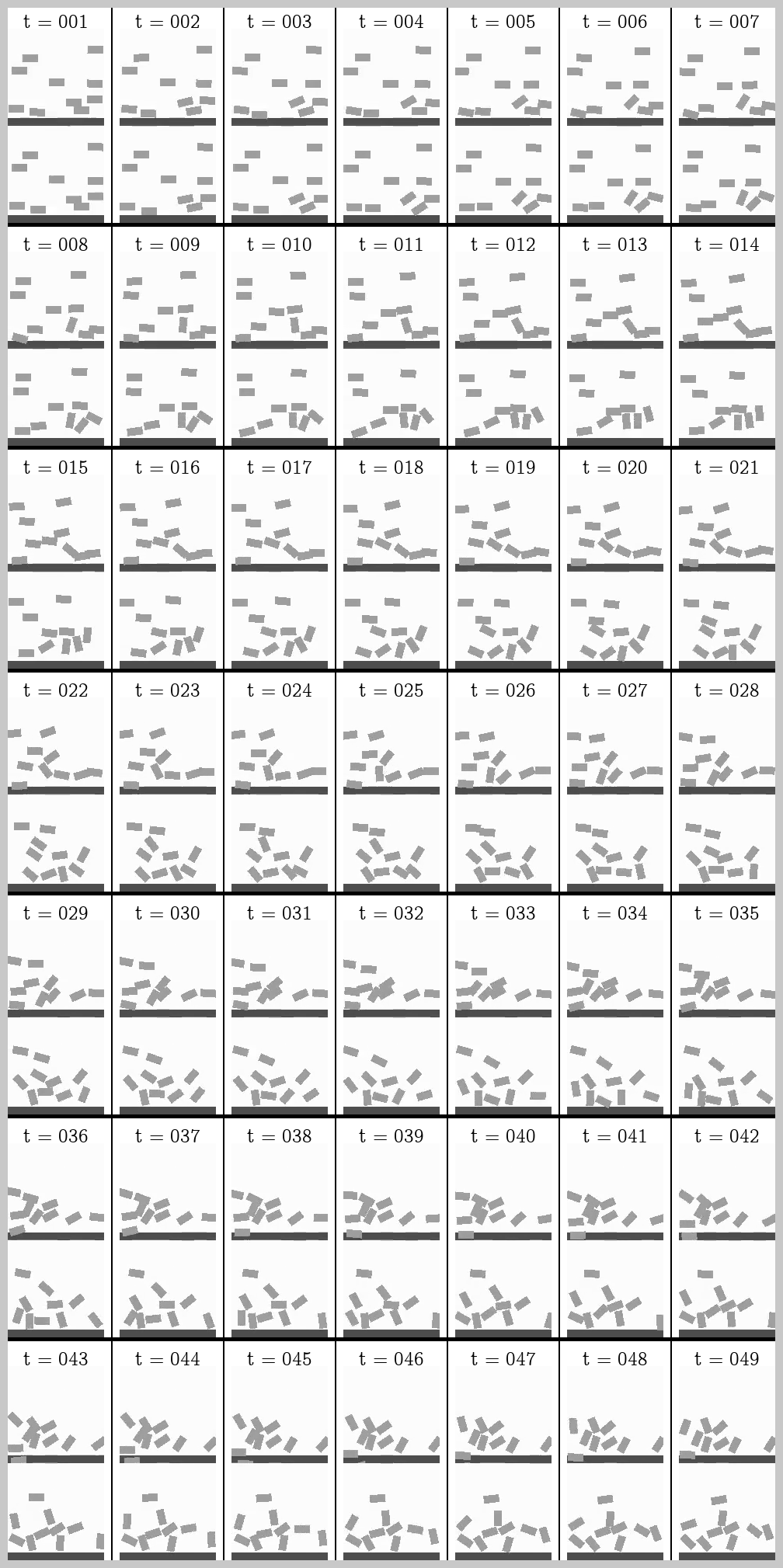}
    \caption{Example rollout using \texttt{Equi-CliffordTransformer}, trained on 1000 episodes of 10 rectangles using sequence length 2. For each timestep, model prediction is shown in the top panel, and the ground truth environment simulation in the bottom panel.}
    \label{fig:supp_rollout_equi_cliffordtransformer_10rect_2seq}
\end{figure}

\begin{figure}
    \centering
    \includegraphics[width=0.8\linewidth]{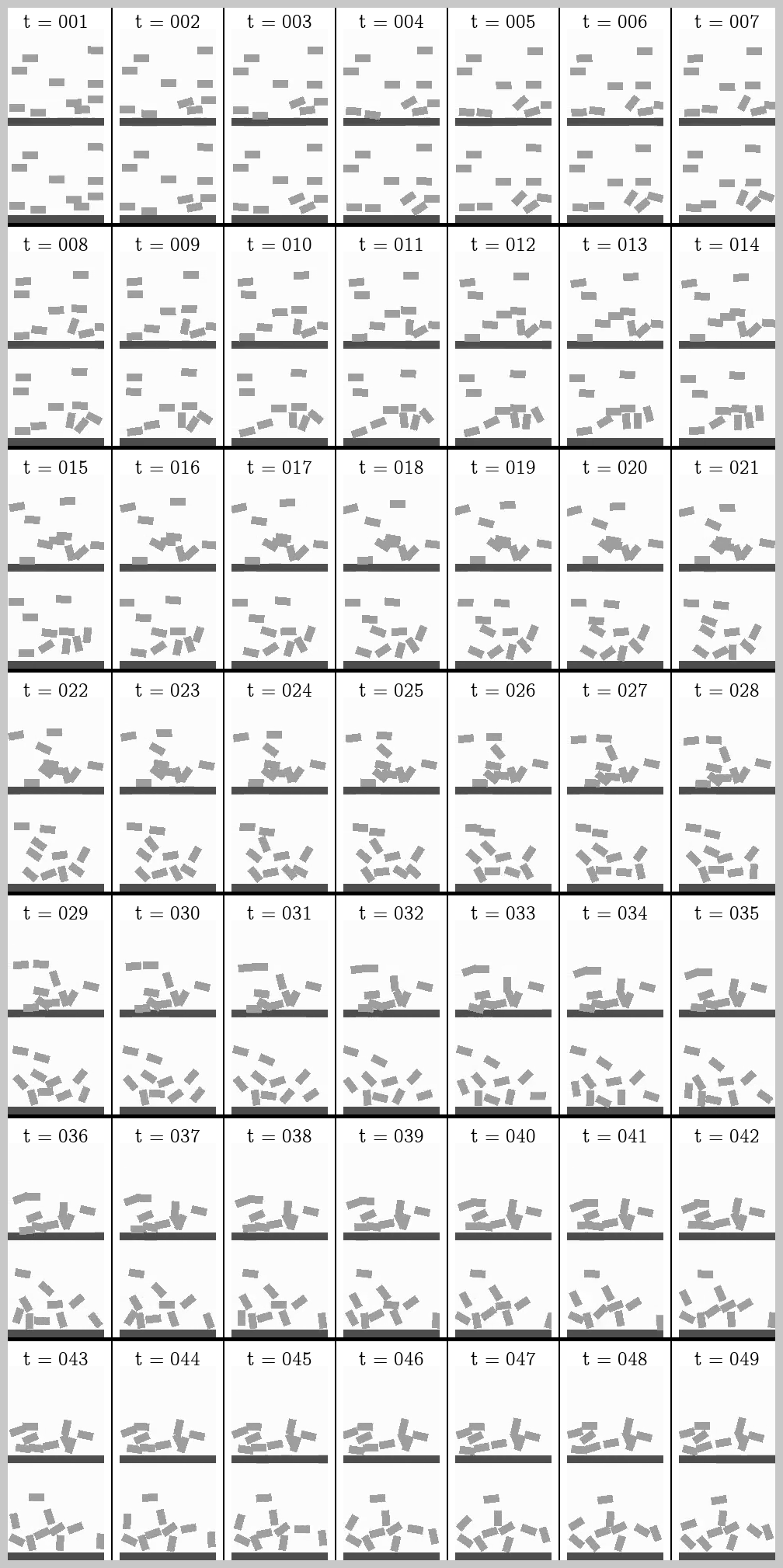}
    \caption{Example rollout using \texttt{S-CliffordTransformer}, trained on 1000 episodes of 10 rectangles using sequence length 2. For each timestep, model prediction is shown in the top panel, and the ground truth environment simulation in the bottom panel.}
    \label{fig:supp_rollout_cliffordtransformer_10rect_2seq}
\end{figure}

\begin{figure}
    \centering
    \includegraphics[width=0.80\linewidth]{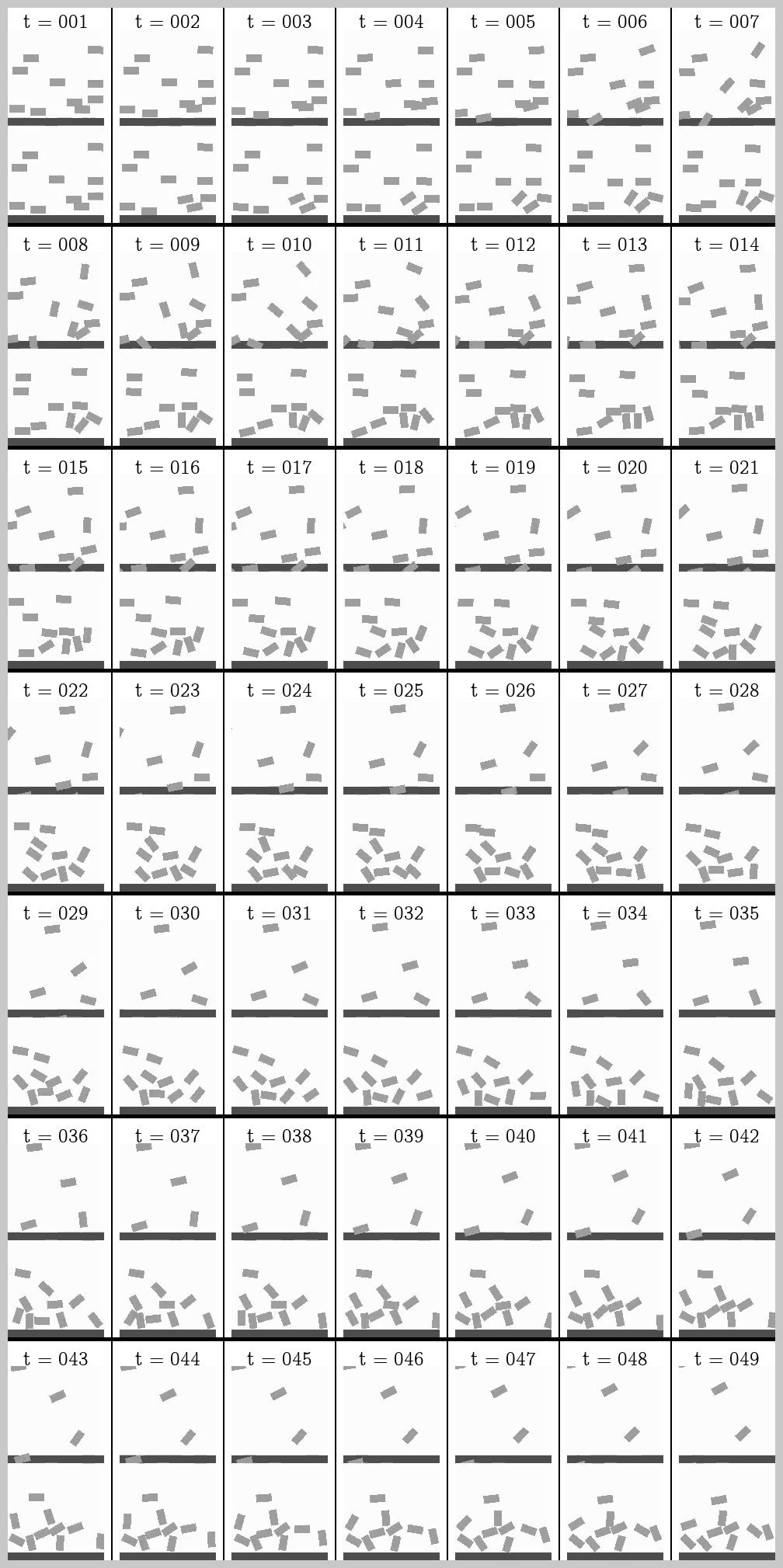}
    \caption{Example rollout using the baseline \texttt{Transformer}, trained on 1000 episodes of 10 rectangles using sequence length 2. For each timestep, model prediction is shown in the top panel, and the ground truth environment simulation in the bottom panel.}
    \label{fig:supp_rollout_transformer_10rect_2seq}
\end{figure}

Finally, for the more challenging task of predicting the dynamics of 10 rectangles, we compare \texttt{E-CliffordTransformer}, \texttt{S-CliffordTransformer} and baseline \texttt{Transformer} in Figures \ref{fig:supp_rollout_equi_cliffordtransformer_10rect_2seq}, \ref{fig:supp_rollout_cliffordtransformer_10rect_2seq}, \ref{fig:supp_rollout_transformer_10rect_2seq}. Our \texttt{S-CliffordTransformer} in Figure \ref{fig:supp_rollout_cliffordtransformer_10rect_2seq} shows rough qualitative agreement with ground truth up to timestep 28, but the predicted objects tend to dissipate energy more quickly than their ground truth counterparts. The \texttt{Equi-CliffordTransformer} also shows rough qualitative agreement, but note that sometimes there are predicted penetrations between dynamic and static objects in early timesteps such as in timestep 8. The baseline transformer in Figure \ref{fig:supp_rollout_transformer_10rect_2seq} quickly diverges, failing to restrict the dynamic objects within the confines of the learned static environment.

\end{document}